\documentclass[pdflatex,sn-mathphys-num]{sn-jnl}

\usepackage{graphicx}%
\usepackage{multirow}%
\usepackage{amsmath,amssymb,amsfonts}%
\usepackage{amsthm}%
\usepackage{mathrsfs}%
\usepackage[title]{appendix}%
\usepackage{xcolor}%
\usepackage{textcomp}%
\usepackage{manyfoot}%
\usepackage{booktabs}%
\usepackage{algorithm}%
\usepackage{algorithmic}%
\usepackage{listings}%
\usepackage{rotating}
\usepackage{array}

\theoremstyle{thmstyleone}%
\theoremstyle{thmstyletwo}%

\theoremstyle{thmstylethree}%

\begin{document}

\title[Article Title]{ASGMamba: Adaptive Spectral Gating Mamba for Multivariate Time Series Forecasting}


\author[1]{\fnm{Qianyang} \sur{Li}}\email{liqianyang@stu.xjtu.edu.cn}
\author*[1]{\fnm{Xingjun} \sur{Zhang}}\email{xjzhang@xjtu.edu.cn}
\author[1]{\fnm{Shaoxun} \sur{Wang}}\email{shaoxunwang@stu.xjtu.edu.cn}
\author[2]{\fnm{Jia} \sur{Wei}}\email{weijia4473@mail.tsinghua.edu.cn}
\author[3]{\fnm{Yueqi} \sur{Xing}}\email{xingyueqi@newbeiyang.com}
\affil*[1]{\orgdiv{School of Computer Science and Technology}, \orgname{Xi'an Jiaotong University}, \orgaddress{ \city{Xi’an} \postcode{710049}, \country{China}}}

\affil[2]{\orgdiv{Department of Computer Science and Technology}, \orgname{Tsinghua University},  \orgaddress{ \city{Beijing} \postcode{100084}, \country{China}}}
\affil[3]{\orgdiv{Department of Research and Development}, \orgname{Shandong New Beiyang Information Technology Co., Ltd}, \orgaddress{ \city{Weihai} \postcode{264200}, \country{China}}}

\abstract{
Long-term multivariate time series forecasting (LTSF) plays a crucial role in various high-performance computing applications, including real-time energy grid management and large-scale traffic flow simulation. However, existing solutions face a dilemma: Transformer-based models suffer from quadratic complexity, limiting their scalability on long sequences, while linear State Space Models (SSMs) often struggle to distinguish valuable signals from high-frequency noise, leading to wasted state capacity. To bridge this gap, we propose ASGMamba, an efficient forecasting framework designed for resource-constrained supercomputing environments. ASGMamba integrates a lightweight Adaptive Spectral Gating (ASG) mechanism that dynamically filters noise based on local spectral energy, enabling the Mamba backbone to focus its state evolution on robust temporal dynamics. Furthermore, we introduce a hierarchical multi-scale architecture with variable-specific Node Embeddings to capture diverse physical characteristics. Extensive experiments on nine benchmarks demonstrate that ASGMamba achieves state-of-the-art accuracy. While keeping strictly $$\mathcal{O}(L)$$ complexity we significantly reduce the memory usage on long-horizon tasks, thus establishing ASGMamba as a scalable solution for high-throughput forecasting in resource limited environments.The code is available at \url{https://github.com/hit636/ASGMamba}}




\maketitle

\section{Introduction}
\label{intro}
%
%
%
%

Long-term multivariate time series forecasting (LTSF) is essential in applications such as decision-making in complex systems \cite{liu2023largest}, ranging from high-frequency financial markets \cite{sezer2020financial,DBLP:journals/csur/ArsenaultWP25} to real-time energy grid management \cite{deb2017review,LAGO2021116983}. 
With the exponential growth of sensor data, deploying accurate forecasting models on high-performance computing (HPC) infrastructures has become a critical challenge. In these operational environments, a model is defined not only by its statistical accuracy but also by its computational efficiency \cite{wang2024deep,DBLP:journals/tjs/YuYWSFL25}. 
Real-world data streams typically exhibit entangled multi-scale dynamics, where critical long-term trends are obscured by high-frequency stochastic noise \cite{kim2025comprehensive, qiu2024tfb}. 
Consequently, the central challenge in LTSF is to distinguish structurally informative signals from noise and capture long-range dependencies without incurring prohibitive memory or latency costs.

In recent years, Transformer-based architectures have dominated LTSF \cite{nie2022time,liu2023itransformer, DBLP:journals/csr/ZhaoCXCWB26}. 
While leveraging global self-attention enables effective modeling of historical dependencies, these models suffer from a quadratic computational complexity $\mathcal{O}(L^2)$ with respect to the sequence length $L$. 
Beyond theoretical complexity, the heavy memory footprint of attention maps often saturates GPU bandwidth for ultra-long sequences, creating a bottleneck for high-throughput applications. 
Sparse attention variants, such as Informer \cite{zhou2021informer}, attempt to reduce this to $\mathcal{O}(L \log L)$ via heuristic sampling, but often at the cost of discarding fine-grained signal information. 
However, Structured State Space Models (SSMs) \cite{DBLP:journals/corr/abs-2312-00752, aoki2013state} offer a more efficient alternative, enabling linear scaling for long-range temporal dependencies. By framing sequence modeling as a discretized continuous-time system \cite{DBLP:conf/iclr/GuGR22}, Mamba achieves strict linear scaling $\mathcal{O}(L)$ through a hardware-aware parallel scan, offering a rigorous foundation for processing massive temporal contexts.

Although SSMs offer efficiency benefits, applying standard Mamba architectures to noisy time series demonstrates a surprising robustness-efficiency conflict.
The standard selective scan mechanism processes input tokens sequentially in the time domain. 
However, distinguishing valid high-frequency signals from stochastic noise purely in the time domain requires the model to approximate complex filtering operations, which expends significant state capacity. 
Depleted of latent state information, the limited latent state of the SSM is often saturated by high-frequency noise, leaving little capacity left to model the underlying long-term trend \cite{ghil2002advanced}.
We therefore contend this is a state efficiency problem: compelling a linear recurrent system to approximate global spectral filtering from scratch is computationally suboptimal compared to supplying an explicit, lightweight spectral prior.

Furthermore, to preserve linear complexity, many architectures decouple multivariate dependencies via a Channel-Independent (CI) protocol \cite{DBLP:journals/corr/abs-2502-10721}. Although efficient, this strategy inherently discards semantic context, as the model learns a shared representation for all variates regardless of their physical nature. Treating heterogeneous signals like voltage and temperature with identical dynamics inevitably degrades performance when their underlying spectral properties differ significantly.

To resolve these conflicts, we propose ASGMamba (Adaptive Spectral Gating Mamba), a linear-complexity framework that conditions state evolution on spectral energy distribution. 
Many existing methods either forecast directly in the frequency domain or apply global Fourier transforms \cite{zhou2022fedformer,DBLP:journals/tjs/ChenYZH25} with $\mathcal{O}(L \log L)$ cost, breaking support for streaming prediction. Different from these methods, ASGMamba follows the local frequency analysis to condition adaptive gating.
From a system viewpoint, our Adaptive Spectral Gating (ASG) is a data-dependent frequency-selective filter that suppresses noise-dominated components at the input stage with local patch energy information, thereby forcing Mamba backbone to reserve its state capacity for robust dynamics.
Importantly, by applying FFT on fixed-size patches, we strictly maintain $\mathcal{O}(L)$ complexity.
Furthermore, we employ learnable Node Embeddings to recover variable-specific semantics in the shared backbone without resorting to quadratic channel mixing.

The main contributions of this work are summarized as follows:

\begin{itemize}
    \item \textbf{Spectral-Conditioned State Evolution Framework:} We propose ASGMamba, which bridges the gap between spectral analysis and linear recurrence. By conditioning the SSM input on local spectral energy density, we prevent high-frequency noise from contaminating the latent state, significantly enhancing the effective capacity of the model for trend modeling.
    
    \item \textbf{Input-Dependent Linear Filtering Mechanism:} We introduce the Adaptive Spectral Gating (ASG) module. Functioning as a learnable filter, it dynamically modulates input fidelity based on frequency properties. This acts as a computation-efficient prior that effectively separates signal from noise within a strict $\mathcal{O}(L)$ budget.
    
    \item \textbf{Hierarchical Multi-Scale System Design:} We implement a multi-branch architecture augmented with learnable Node Embeddings. This design enables the model to capture multi-granularity temporal patterns and distinct physical semantics of variables simultaneously, ensuring scalability for high-dimensional multivariate data.
    
    \item \textbf{Efficiency-Accuracy Trade-off:} Extensive evaluations on nine real-world benchmarks demonstrate that ASGMamba achieves competitive or superior performance compared to state-of-the-art Transformer and SSM baselines. Notably, it exhibits significantly reduced memory footprint and inference latency in long-sequence scenarios, validating its suitability for resource-constrained supercomputing environments.
\end{itemize}

The remainder of this paper is organized as follows. Section \ref{sec:related} reviews related work. Section \ref{sec:preliminaries} shows the preliminaries. Section \ref{sec:method} details the architecture of ASGMamba. Section \ref{sec:exp} presents our experimental setup and results. Finally, Section \ref{sec:con} concludes the paper.

\section{Related Work}
\label{sec:related}

This section reviews literature closely related to our work, organized into three research streams: deep learning methods for long-term time series forecasting (LTSF), state space models (SSMs) for efficient sequence modeling, and frequency-domain analysis in deep neural networks.

\subsection{Deep Learning Methods for Time Series Forecasting}
Most data-driven forecasting approaches prior to deep learning were statistical models such as ARIMA and Prophet \cite{taylor2018forecasting}, which adopt linear dynamics assumption and are hard to model the nonlinear patterns frequently presented in large-scale sensor data.
When deep learning emerges, RNNs, LSTMs and GRUs become widely used deep models for sequential modeling due to their recurrent property.
However, their recurrent nature makes them hard to be parallelized and thus becomes inefficient to model ultra-long sequences, especially in high sampling frequency scenarios \cite{kim2025comprehensive}.
Transformer \cite{vaswani2017attention} leverages self-attention to model global dependency in sequence without recurrent property.
Follow-up methods like LogTrans \cite{nie2022time} and Informer \cite{zhou2021informer} also reduce the quadratic complexity of attention by leveraging sparse attention mechanism and further apply decomposition strategy to model trend and seasonal components independently \cite{zhou2022fedformer,wu2021autoformer}.
Patch-based methods like PatchTST \cite{nie2022time} aggregate local temporal segments into tokens to enhance local pattern modeling and reduce sequence length.
Proposed as an efficient MLP-based solution, the MSTF \cite{DBLP:journals/tjs/ZhouJLCZ25} model enhances long-term time series forecasting by combining a Time Reverse and Transform block for global perception with a Dynamic Combination Reconstruction block to effectively capture multi-scale temporal dependencies.
Nevertheless, these Transformer-based models still adopt attention mechanism whose computational cost and memory cost both increase superlinearly with sequence length, which motivates us to explore more scalable architectures for long-horizon forecasting.

\subsection{State Space Models and Mamba}
State Space Models (SSMs) have recently re-emerged as an efficient alternative to attention-based architectures, aiming to combine parallelizable training with fast autoregressive inference.
The Structured State Space model (S4) \cite{somvanshi2025s4mambacomprehensivesurvey} demonstrated that imposing structured parameterizations on state transitions enables SSMs to model long-range dependencies with linear complexity.
Building upon S4, Mamba \cite{DBLP:journals/corr/abs-2312-00752} introduced a selective scanning mechanism that allows model parameters to be input-dependent, achieving strong performance in large-scale sequence modeling tasks while preserving linear scaling.

In the context of time series forecasting, recent studies have explored adapting SSMs to multivariate and long-horizon settings.
For example, S4ND \cite{DBLP:conf/nips/NguyenGGDSDBR22} extended S4 to multidimensional signals, and TimeMachine \cite{DBLP:conf/ecai/AhamedC24} applied Mamba-style architectures to forecasting tasks.
However, these approaches primarily operate in the time domain and do not explicitly account for the spectral characteristics of sensor data.
As a result, high-frequency noise and informative periodic components are processed within a shared state space, potentially limiting robustness in highly volatile environments.

\subsection{Frequency-Domain Analysis in Deep Learning}
Frequency-domain analysis has been applied in many fields to describe periodicity and filter out noise from time series.
In deep learning, FEDformer \cite{zhou2022fedformer} applied Fourier transforms in combination with attention mechanisms to model global dependencies more efficiently. TimesNet \cite{DBLP:conf/iclr/WuHLZ0L23} represented time series in a two-dimensional space according to the dominant periods, and FITS \cite{DBLP:conf/iclr/XuZ024} applied forecasting in the complex frequency domain. Recently, MSFMoE \cite{DBLP:journals/ijon/BaoTJH26} applied multi-scale frequency filtering in combination with a mixture-of-experts architecture to improve long-term forecasting performance. 

Although these frequency-based methods have achieved promising performance, most of them apply spectral transformations in combination with global feature extraction or static filtering. That is, they transfer the entire sequence using fixed mappings and do not modulate the information flow via finer time grains in an adaptive way. Therefore, we attempt to explore how the spectral information can be applied as a structural gating signal in an SSM framework.

\section{Preliminaries}
\label{sec:preliminaries}

In this section, we formulate the forecasting problem and provide the theoretical background of the Structured State Space Model (SSM), which serves as the backbone of our efficient framework.

\subsection{Problem Formulation}
The task of Multivariate Time Series (MTS) forecasting aims to predict future trajectories based on historical observations. 
Let $\mathcal{D} = \{(\mathbf{X}^{(i)}, \mathbf{Y}^{(i)})\}_{i=1}^{\mathcal{N}}$ denote a dataset of time series samples.
For a single instance $i$, $\mathbf{X}^{(i)} = \{\mathbf{x}_1, \dots, \mathbf{x}_L\} \in \mathbb{R}^{L \times M}$ represents the historical look-back sequence, where $L$ is the observation window length and $M$ is the number of variates (channels).
The objective is to predict the future sequence $\mathbf{Y}^{(i)} = \{\mathbf{x}_{L+1}, \dots, \mathbf{x}_{L+T}\} \in \mathbb{R}^{T \times M}$ over a horizon $T$.

Formally, we seek to learn a mapping function $\mathcal{F}_\theta: \mathbb{R}^{L \times M} \to \mathbb{R}^{T \times M}$, parameterized by $\theta$, that minimizes the prediction discrepancy on unseen data. To leverage parallel computing hardware, the model processes inputs in mini-batches. 
Consequently, the input to the algorithm is represented as a tensor $\mathbf{X}_{in} \in \mathbb{R}^{B \times L \times M}$, where $B$ denotes the batch size. 
Furthermore, under the Channel-Independent (CI) strategy, this batch is reshaped to $\mathbb{R}^{(B \cdot M) \times L \times 1}$ to share backbone parameters across all variates efficiently.

\subsection{Efficient Modeling via State Space Models}
The core efficiency of ASGMamba stems from the Structured State Space Model (SSM) \cite{somvanshi2025s4mambacomprehensivesurvey}. SSMs map a 1D input stimulation $x(t) \in \mathbb{R}$ to an output $y(t) \in \mathbb{R}$ through a latent state $h(t) \in \mathbb{R}^N$.
The system is modeled as a linear Time-Invariant (LTI) continuous system:
\begin{equation}
    \label{eq:ssm_continuous}
    \dot{h}(t) = \mathbf{A}h(t) + \mathbf{B}x(t), \quad y(t) = \mathbf{C}h(t),
\end{equation}
where $\mathbf{A} \in \mathbb{R}^{N \times N}$ is the evolution matrix, and $\mathbf{B}, \mathbf{C}$ are projection parameters. To handle discrete time series, the system is discretized using a time scale parameter $\Delta$, transforming Eq. \eqref{eq:ssm_continuous} into a recurrence relation:
\begin{equation}
    \label{eq:ssm_discrete}
    h_t = \bar{\mathbf{A}}h_{t-1} + \bar{\mathbf{B}}x_t, \quad y_t = \mathbf{C}h_t.
\end{equation}
This recursive form allows for fast autoregressive inference ($O(1)$ per step). Alternatively, the system can be computed via global convolution during training, enabling parallelization similar to Transformers.

In State Space Models (SSMs), all parameters are typically fixed. In contrast, the Mamba \cite{DBLP:journals/corr/abs-2312-00752} model makes the parameter set \(\mathbf{B}, \mathbf{C}, \Delta\) functions of \(x_t\). Hence, Mamba can decide at each time step which parameters to propagate or forget along the sequence (e.g., skip connections for zeroing out noise).
Despite the input-dependence, Mamba is still able to evaluate Eq. \eqref{eq:ssm_discrete} efficiently via a convolution-like operation using a hardware-aware parallel scan algorithm.
More importantly, due to its linear computational complexity \(O(L)\), this architecture is significantly more efficient than the quadratic \(O(L^2)\) attention mechanism used in Transformers when processing long historical  sequences.

\section{Proposed Method}
\label{sec:method}

This section presents the architecture of ASGMnable (Adaptive Spectral Gating Mamba). Moreover, the overarching design objective may suggest that spectral-conditioned state evolution should be enforced within a strictly linear computational budget. Given that standard State Space Models (SSMs) provide efficient $\mathcal{O}(L)$ inference, the time-domain selective scan mechanism appears inefficient at filtering broadband noise. However, noise may lead to state saturation. To resolve this, ASGMamba injects a lightweight spectral prior directly into the backbone, functioning as an input-modulated filter that disentangles valid signals from noise prior to state encoding.
\begin{figure*}[t]
    \centering
    \includegraphics[width=\linewidth]{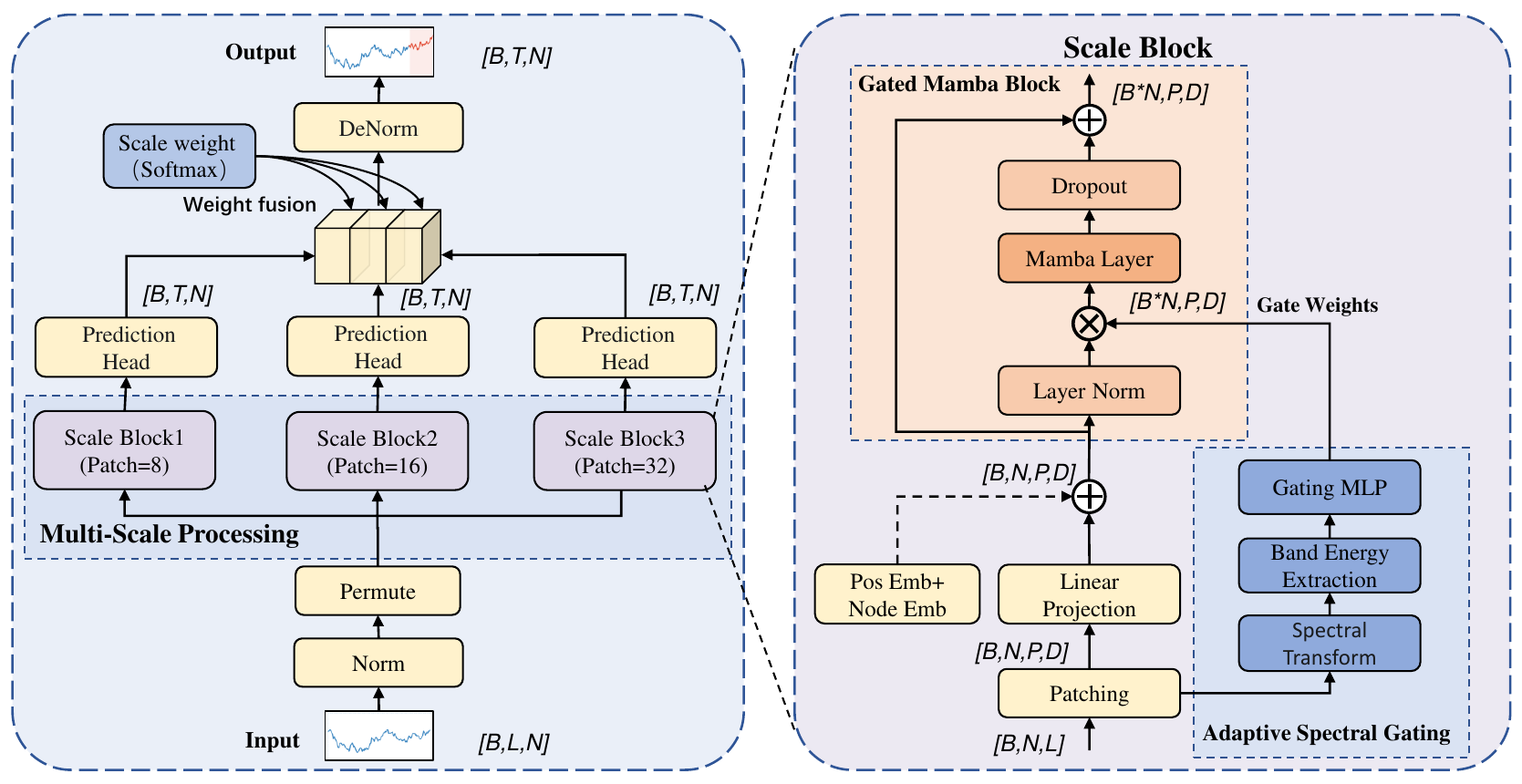} 
    \caption{System Architecture of ASGMamba. The framework employs a multi-scale parallel design ($P=\{8, 16, 32\}$). The core innovation is the \textbf{Adaptive Spectral Gating (ASG)} module within the Scale Block. By computing local spectral energy density via patch-level FFT, ASG generates a data-dependent gate $\mathbf{G}$. This gate acts as a frequency-selective filter, modulating the input $\mathbf{U}$ before it enters the Mamba encoder, thereby preventing high-frequency noise from contaminating the latent state space.}
    \label{fig:architecture}
\end{figure*}

As illustrated in Fig. \ref{fig:architecture}, the model employs a multi-level approach with adaptive patching strategies to preserve the identity of the data.
To handle non-stationarity in raw sensor data, we apply Reversible Instance Normalization (RevIN)~\cite{kim2021reversible} at the instance level.
Following normalization, we adopt a Channel-Independent (CI) strategy, reshaping the input to $\tilde{\mathcal{X}} \in \mathbb{R}^{(B \cdot N) \times L \times 1}$. 
This treats all $N$ variates as independent sequences, reducing parameter complexity from $\mathcal{O}(N^2)$ to $\mathcal{O}(1)$.
To reduce confusion in the Channel-Independent approach, we introduce learnable node embeddings to better handle different variable types. The algorithmic workflow is formally detailed in Alg.\ref{alg:ASGMamba}.

\begin{algorithm}[t]
\caption{Forward Pass of ASGMamba}
\label{alg:ASGMamba}
\begin{algorithmic}[1]
\renewcommand{\algorithmicrequire}{\textbf{Input:}}
\renewcommand{\algorithmicensure}{\textbf{Output:}}

\REQUIRE Batch input $\mathbf{X}_{in} \in \mathbb{R}^{B \times L \times N}$
\ENSURE Forecast $\mathbf{Y}_{out} \in \mathbb{R}^{B \times T \times N}$

\vspace{0.3em}
\STATE \textbf{Preprocessing:}
\STATE $\mathbf{X} \leftarrow \text{RevIN}(\mathbf{X}_{in}^{\top}) \in \mathbb{R}^{B \times V \times N}$
\STATE $\mathcal{Y}_{set} \leftarrow \emptyset$

\vspace{0.3em}
\FOR{patch size $P \in \{8, 16, 32\}$}
    \STATE \textit{// Patching and Context Injection}
    \STATE $\mathbf{X}_{p} \leftarrow \text{OverlappingPatching}(\mathbf{X}, P)$
    \STATE $\mathbf{U} \leftarrow \text{PatchEmb}(\mathbf{X}_{p}) + \mathbf{E}_{pos} + \mathbf{E}_{node}$

    \vspace{0.3em}
    \STATE \textit{// Adaptive Spectral Gating}
    \STATE $\mathbf{F} \leftarrow |\text{rFFT}(\mathbf{X}_{p})|^2$ \COMMENT{Local Spectral Density}
    \STATE $\mathbf{v}_{spec} \leftarrow \text{BandAgg}(\mathbf{F})$ \COMMENT{Low/Mid/High Energy}
    \STATE $\mathbf{G} \leftarrow \text{Sigmoid}(\text{MLP}(\mathbf{v}_{spec}))$
    
    \vspace{0.3em}
    \STATE \textit{// Spectral-Conditioned SSM}
    \STATE $\mathbf{U}_{gated} \leftarrow \mathbf{U} \odot \mathbf{G}$ \COMMENT{Noise Suppression}
    \STATE $\mathbf{H} \leftarrow \text{Mamba}(\mathbf{U}_{gated}) + \mathbf{U}_{gated}$

    \vspace{0.3em}
    \STATE \textit{// Projection}
    \STATE $\mathbf{Y}^{(P)} \leftarrow \text{Head}(\mathbf{H})$
    \STATE $\mathcal{Y}_{set} \leftarrow \mathcal{Y}_{set} \cup \{\mathbf{Y}^{(P)}\}$
\ENDFOR

\vspace{0.3em}
\STATE \textbf{Scale Fusion \& Output:}
\STATE $\mathbf{Y}_{fused} \leftarrow \sum_{k} \text{Softmax}(\mathbf{w}_{scale})_k \cdot \mathbf{Y}^{(k)}$
\STATE $\mathbf{Y}_{out} \leftarrow \text{RevIN}^{-1}(\mathbf{Y}_{fused})^{\top}$

\RETURN $\mathbf{Y}_{out}$
\end{algorithmic}
\end{algorithm}

\subsection{Multi-Scale Patching with Identity Preservation}
\label{subsec:patching}

Real-world systems evolve across diverse temporal granularities. To align the model's receptive field with these physical scales, ASGMamba employs a multi-branch architecture with $K=3$ branches, corresponding to patch sizes $P_k \in \{8, 16, 32\}$. This design is motivated by the observation that different spectral bands require different effective temporal resolutions.

\subsubsection{Overlapping Patching Process}
Standard non-overlapping patching creates hard boundaries that can sever local dependencies. From a signal processing perspective, such discontinuities introduce high-frequency artifacts (spectral leakage) when applying FFT. 
To mitigate this, we adopt an overlapping strategy with stride $S_k = \frac{P_k}{2}$ (50\% overlap).
This redundancy ensures that boundary information in patch $i$ becomes central information in patch $i+1$, acting as a smoothing regularizer that preserves spectral continuity for the subsequent gating module.
The input sequence is unfolded into $N_k$ patches, yielding a tensor $\mathbf{X}_{patch}^{(k)} \in \mathbb{R}^{(B \cdot N) \times N_k \times P_k}$, which is then projected to a $D$-dimensional latent space $\mathbf{Z}^{(k)}_{raw}$.

\subsubsection{Injecting Position and Variable Identity}
To compensate for the information loss caused by the CI strategy, we inject explicit semantic identifiers:
\begin{itemize}
    \item \textbf{Positional Embedding ($\mathbf{E}_{pos}$):} A learnable tensor informs the SSM of global temporal ordering.
    \item \textbf{Node Embedding ($\mathbf{E}_{node}$):} 
    To resolve the identity crisis of CI models, we introduce a learnable matrix $\mathbf{E}_{node} \in \mathbb{R}^{M \times D}$. The vector $\mathbf{e}_m = \mathbf{E}_{node}[m]$ serves as a static semantic descriptor for variate $m$. 
    By adding this embedding to the input, we condition the shared Mamba backbone to adapt its state dynamics to the specific physical properties of each variable (e.g., distinct periodicities of voltage vs. load) without the quadratic cost of inter-channel attention.
\end{itemize}
The final context-aware input is formulated as:
\begin{equation}
    \mathbf{Z}^{(k)}_0 = \mathbf{Z}^{(k)}_{raw} + \mathbf{E}_{pos} + \mathbf{E}_{node}.
\end{equation}

\subsection{Adaptive Spectral Gating Mamba Layer}
\label{subsec:ASGMamba_layer}

This module constitutes the core contribution of our framework. It addresses the limitation of standard SSMs, which process all input tokens with equal weight, leading to efficient but indiscriminate modeling. We introduce a spectral filtering stage prior to the state space equation.

\subsubsection{Adaptive Spectral Gating (ASG)}
The Spectral Gating mechanism acts as a filter that adjusts based on the input data, allowing the model to focus on relevant frequencies. It assesses the spectral quality of each local patch and attenuates noise-dominated segments before they are encoded into the latent state.

\paragraph{Local Spectral Transformation.}
We apply a real-valued Fast Fourier Transform (rFFT) along the temporal dimension of each patch $\mathbf{X}^{(k)}_{\text{patch}}$:
\begin{equation}
\mathcal{F}_{\text{patch}} = \text{rFFT}(\mathbf{X}^{(k)}_{\text{patch}})
\in \mathbb{C}^{(B \cdot N) \times N_k \times (\lfloor P_k/2 \rfloor + 1)}.
\end{equation}
Instead of applying global Fourier Transforms that scale at $\mathcal{O}(L \log L)$ and disrupt continuous data processing, the ASGMamba model incorporates a lightweight spectral gating mechanism, focusing on local operations to improve efficiency.

\paragraph{Band-wise Energy Aggregation.}
To obtain a robust spectral descriptor, we compute the spectral power density $S = |\mathcal{F}_{\text{patch}}|^2$ and aggregate it into three coarse frequency bands. This partition is motivated by empirical observations of time series spectral concentration:
\begin{itemize}
    \item \textbf{Low Band ($[0, \frac{1}{3}f_{\text{Nyq}}]$):} Captures long-term trends and DC components.
    \item \textbf{Mid Band ($(\frac{1}{3}, \frac{2}{3}]f_{\text{Nyq}}$):} Encodes dominant periodicities and seasonality.
    \item \textbf{High Band ($(\frac{2}{3}, 1]f_{\text{Nyq}}$):} Corresponds to rapid fluctuations, often dominated by sensor noise or stochastic jitter.
\end{itemize}
The spectral descriptor $\mathbf{v}_{\text{spec}} \in \mathbb{R}^3$ is derived by summing normalized energy within these bands. This low-dimensional representation allows the gating mechanism to reason about signal composition (e.g., "noisy" vs. "trend-driven") without overfitting to specific frequency bins.

\paragraph{Gate Generation.}
A lightweight MLP maps the spectral descriptor to a gating tensor:
\begin{equation}
\mathbf{G} = \sigma\!\left( \mathbf{W}_{g2}\,
\text{ReLU}(\mathbf{W}_{g1}\mathbf{v}_{\text{spec}}) \right),
\end{equation}
where $\sigma(\cdot)$ is the Sigmoid function. The gate $\mathbf{G} \in (0,1)^{(B \cdot M) \times N_k \times D}$ effectively acts as a confidence score: patches with high noise energy (High Band dominance) result in lower gate values ($\mathbf{G} \to 0$), while trend-rich patches are preserved ($\mathbf{G} \to 1$).

\subsubsection{Spectral-Conditioned Mamba Block}
We integrate the spectral gate into a residual Mamba block. Let $\mathbf{Z}_{\text{in}}$ be the input embedding. The gated representation is computed as:
\begin{equation}
\mathbf{Z}_{\text{gated}} = \text{LayerNorm}(\mathbf{Z}_{\text{in}}) \odot \mathbf{G}.
\end{equation}
This modulated signal drives the selective State Space Model (SSM):
\begin{equation}
\mathbf{H}_t = \mathbf{A}\mathbf{H}_{t-1} + \mathbf{B}\mathbf{Z}_{\text{gated},t},
\quad
\mathbf{Y}_t = \mathbf{C}\mathbf{H}_t.
\end{equation}
\textbf{System Interpretation:} By attenuating $\mathbf{Z}_{\text{gated}}$ via $\mathbf{G}$, we effectively reduce the magnitude of the input projection $\mathbf{B}\mathbf{x}_t$ for noise-dominated tokens. This prevents the recurrent state $\mathbf{H}_t$ from updating its dynamics based on spurious fluctuations, thereby conserving state capacity for valid long-term dependencies.
The block concludes with a residual connection to facilitate gradient flow: $\mathbf{Z}_{\text{out}} = \text{Dropout}(\text{Mamba}(\mathbf{Z}_{\text{gated}})) + \mathbf{Z}_{\text{in}}$.

\subsection{Adaptive Multi-Scale Fusion}
\label{subsec:fusion}

To accommodate variables with diverse temporal dynamics (e.g., slow-moving trends vs. rapid cycles), we fuse predictions from the three parallel branches ($P \in \{8, 16, 32\}$).
Let $\hat{\mathbf{Y}}_k$ be the forecast from branch $k$. We employ a learnable convex combination parameterized by $\mathbf{w}_{scale} \in \mathbb{R}^3$:
\begin{equation}
    \hat{\mathbf{Y}}_{fused} = \sum_{k=1}^{3} \frac{\exp(\mathbf{w}_{scale}^{(k)})}{\sum_{j=1}^3 \exp(\mathbf{w}_{scale}^{(j)})} \cdot \hat{\mathbf{Y}}_k.
\end{equation}
This mechanism allows the model to dynamically prioritize the temporal resolution that best fits the data's inherent frequency characteristics (e.g., favoring larger patches for trend-dominated series).

\subsection{Complexity and Scalability Analysis}
\label{subsec:complexity}

A critical requirement for deploying forecasting models in supercomputing environments is the ability to scale efficiently with sequence length $L$. Here, we formally derive the computational complexity of ASGMamba.

\textbf{Linear Complexity of Spectral Gating.} 
Standard global Frequency-domain methods (e.g., FEDformer) rely on global FFT, incurring a complexity of $\mathcal{O}(L \log L)$. In contrast, ASGMamba applies FFT on local patches of a fixed size $P$ (e.g., $P=16$).
Let $N \approx L/S$ be the number of patches, where $S$ is the stride. The total complexity for the gating mechanism is:
\begin{equation}
    \mathcal{C}_{gate} = N \times \mathcal{O}(P \log P) \approx \frac{L}{S} \cdot P \log P.
\end{equation}
Since the patch size $P$ and stride $S$ are small constants ($P \ll L$), the term $\frac{P \log P}{S}$ acts as a constant multiplier. Consequently, the asymptotic complexity is strictly linear with respect to the sequence length, i.e., $\mathcal{O}(L)$.

\textbf{Implementation Details.} 
The Adaptive Spectral Gating (ASG) module is implemented with high computational efficiency. 
(1) \textbf{Frequency Bands}: We partition the spectrum into $K_{freq}=3$ equal bands (Low, Mid, High) based on the Nyquist frequency to capture trend, periodicity, and noise, respectively.
(2) \textbf{Gating MLP}: The spectral energy vector is processed by a 2-layer MLP with a bottleneck structure. The hidden dimension is set to $D/4$ (reduction ratio $r=4$) with ReLU activation, followed by a Sigmoid function to output valid gate weights in $(0, 1)$.
(3) \textbf{Memory Efficiency}: Combined with the Mamba backbone's linear scaling $\mathcal{O}(L)$, ASGMamba avoids the quadratic memory matrix $\mathcal{O}(L^2)$ of Transformers, enabling high-throughput inference on GPU clusters for ultra-long sequences ($L > 10^3$).

\section{Experiments}
\label{sec:exp}

In this section, we systematically study ASGMamba. First, we evaluate ASGMamba on nine public datasets and demonstrate its superior forecasting accuracy compared to state-of-the-art models. Secondly, we conduct ablative analysis to illustrate the effectiveness of each component. Finally, we analyze the sensitivity of ASGMamba to some critical hyper-parameters.

\subsection{Experimental Setup}
\subsubsection{Datasets} \label{datasets}

In order to thoroughly evaluate our method, we select nine publicly available real-world benchmarks covering various domains, sampling frequencies, and temporal patterns (Statistical information of datasets is provided in Table \ref{tab:dataset_stats}). These nine benchmarks present various forecasting challenges. Specifically, high-dimensional modeling contains Electricity and Traffic which respectively record hourly consumption from hundreds of electric meters and hourly traffic volume on thousands of roads, thus challenging the model to learn fine-grained dynamics and multi-scale seasonality on top of complex spatial-temporal correlations. For fine-grained dynamics, Weather (10-minute intervals) tests model's robustness to nonlinear interactions between variables as well as inherent noises of sensors, and Solar-Energy records power generation of 137 photovoltaic plants, challenging models to model intense volatility from clouds coverage and strict diurnal periodicity. For addressing non-stationarity and volatility, Exchange-Rate shows structural breaks and low signal-to-noise ratio due to global economic conditions, while ETT family (ETTh1, ETTh2, ETTm1, ETTm2) contains electricity consumption measured from transformer at hourly and 15-minute granularity. ETT is more challenging due to its stronger non-stationarity and mixture of regular periodicity and irregular anomalies caused by different operating conditions. In total, these nine benchmarks test models on dependencies, noise, and shifts.

\begin{table}[h!]
\centering
\caption{Statistics of the benchmark datasets used in experiments.}
\label{tab:dataset_stats}
\begin{tabular}{lrrcl}
\toprule
\textbf{Dataset} & \textbf{Variates} & \textbf{Timesteps} & \textbf{Frequency} & \textbf{Domain} \\
\midrule
Electricity      & 321  & 26,304  & 1-hour  & Energy  \\
Traffic          & 862  & 17,520  & 1-hour  & Traffic \\
Weather          & 21   & 52,696  & 10-min  & Weather \\
Exchange-Rate    & 8    & 7,588   & Daily   & Economy \\
Solar            & 137  & 7,176   & 1-hour  & Solar   \\
ETTh1 / ETTh2    & 7    & 17,420  & 1-hour  & Energy  \\
ETTm1 / ETTm2    & 7    & 69,680  & 15-min  & Energy  \\
\bottomrule
\end{tabular}
\end{table}

To evaluate the performance of ASGMamba comprehensively, we select 10 representative state-of-the-art models as baselines. These models encompass the mainstream architectural paradigms discussed in Section \ref{sec:related}, including SSM-based, Transformer-based, and MLP/CNN-based approaches.

\textit{SSM-based Models:}
We compare ASGMamba with the following baselines to demonstrate its architectural benefits over standard State Space Models:

\textbf{S-Mamba} \cite{DBLP:journals/ijon/WangKFWYZWZ25}: A straightforward adaptation of Mamba for time series. It lacks the dynamic spatial reordering feature of our proposal, generally processing variables in a fixed order or independently (Channel-Independent).

\textit{Transformer-based Models:}
These models represent the mainstream SOTA benchmarks, using self-attention to capture long-range temporal relationships:

\textbf{iTransformer} \cite{liu2023itransformer}: It adopts an inverted structure where the whole series of a variate serves as a token. This design enables the attention module to capture multivariate correlations from a global perspective.

\textbf{PatchTST} \cite{nie2022time}: A leading Channel-Independent method that divides series into patches. Although it achieves efficiency and retains local information, it does not account for inter-variable dependencies.

\textbf{Crossformer} \cite{zhang2023crossformer}: It implements a two-stage attention mechanism to capture dependencies in both time and dimension (variate) axes explicitly.

\textbf{FEDformer} \cite{zhou2022fedformer}: By combining seasonal-trend decomposition with Fourier Transform-based frequency attention, it efficiently models global temporal structures.

\textit{MLP- and CNN-based Models:}
This category includes efficient models utilizing inductive biases like decomposition and multi-scale analysis:

\textbf{TimeMixer} \cite{wang2024timemixer}: A multi-scale architecture that decomposes time series into trend and seasonality, handling them via specialized MLP pathways.

\textbf{TimesNet} \cite{DBLP:conf/iclr/WuHLZ0L23}: By reshaping 1D series into 2D tensors based on multiple periods, it employs 2D convolutions to extract intra-period and inter-period features.

\textbf{TiDE} \cite{DBLP:journals/tmlr/DasKLMSY23}
TiDE is an MLP-based encoder-decoder model for long-term forecasting. It encodes historical data and covariates via dense layers, completely avoiding attention mechanisms. Featuring global residual connections, it achieves state-of-the-art accuracy with linear computational complexity, offering significantly faster training and inference than Transformer-based approaches.

 \textbf{DLinear} \cite{zeng2023transformers}: A simple yet effective baseline that decomposes time series into trend and remainder components, processing each with a single linear layer. It challenges the necessity of complex architectures for certain forecasting tasks.

\subsubsection{Implementation Details}
\label{subsubsec:implementation}

Our experimental framework is rigorously designed to ensure reproducibility and to validate the efficiency advantages of the ASGMamba architecture. The implementation details are categorized into three primary dimensions: experimental environment setup, model-specific hyperparameters, and the training optimization protocol.

\noindent\textbf{Experimental Setup.}
All experiments are implemented in PyTorch on a single NVIDIA RTX 4090 (24GB) GPU. Following standard LTSF protocols, we fix the look-back window at $L=96$ and evaluate forecasting performance across four prediction horizons $H \in \{96, 192, 336, 720\}$ on all datasets.

\noindent\textbf{Model and Training Configuration.}
The detailed hyperparameter settings are summarized in Table~\ref{tab:hyperparams}. ASGMamba employs a multi-scale architecture with three parallel branches ($P \in \{8, 16, 32\}$). Crucially, we adopt an overlapping patching strategy with a stride of $S_k = P_k/2$ (50\% overlap) to mitigate boundary artifacts and ensure smoother token transitions. The gating mechanism utilizes a 2-layer bottleneck MLP to map the 3-dimensional spectral energy vector into gating weights.
Training is performed using the Adam optimizer. Due to the rapid convergence of the Mamba backbone, training is capped at 10 epochs. To prevent overfitting, we apply L2 weight decay, dropout, and an early stopping mechanism (patience=5). All results are averaged over five independent runs.

\begin{table}[h]
\centering
\caption{Summary of Hyperparameters for ASGMamba.}
\label{tab:hyperparams}
\begin{tabular}{l|lc}
    \toprule
    \textbf{Category} & \textbf{Parameter} & \textbf{Value} \\
    \midrule
    \multirow{5}{*}{Model Architecture} 
    & Latent Dimension ($D_{model}$) & 128 \\
    & Patch Sizes ($P$) & \{8, 16, 32\} \\
    & Mamba State Dim ($d_{state}$) & 16 \\
    & Mamba Conv Kernel ($d_{conv}$) & 4 \\
    & Expansion Factor ($E$) & 2 \\
    \midrule
    \multirow{5}{*}{Training \& Opt.} 
    & Batch Size & 32 \\
    & Max Epochs & 10 \\
    & Initial Learning Rate & $10^{-3}$ \\
    & L2 Weight Decay & $1 \times 10^{-5}$ \\
    & Dropout Rate & 0.1 \\
    \bottomrule
\end{tabular}
\end{table}
\subsubsection{Evaluation Metrics}
We evaluate our model using two metrics: Mean Squared Error (MSE) and Mean Absolute Error (MAE). Given a prediction $\hat{Y}$ and the corresponding ground truth $Y$ over a horizon of length $H$, these metrics are defined as:
\begin{align}
     \text{MSE} = \frac{1}{H} \sum_{i=1}^{H} (\hat{Y}_i - Y_i)^2 \\
     \text{MAE} = \frac{1}{H} \sum_{i=1}^{H} |\hat{Y}_i - Y_i|
\end{align}
   
For both metrics, lower values indicate better forecasting accuracy.

\begin{sidewaystable*}[p]
\centering
\renewcommand{\arraystretch}{1.4}
\setlength{\tabcolsep}{1.1pt} 
\caption{Long-term forecasting performance comparison with an input length of $L=96$ for prediction horizons $T \in \{96, 192, 336, 720\}$. The best results are in red and the second best are blue. 'Avg' denotes the average performance.}
\resizebox{\linewidth}{!}{%
 \begin{tabular}{l l *{20}{w{r}{3em}}} 
\toprule
\multicolumn{2}{l}{\textbf{Models}} & \multicolumn{2}{c}{\textbf{ASGMamba}} & \multicolumn{2}{c}{\textbf{S-Mamba}} & \multicolumn{2}{c}{\textbf{TimeMixer}} & \multicolumn{2}{c}{\textbf{iTransformer}} & \multicolumn{2}{c}{\textbf{PatchTST}} & \multicolumn{2}{c}{\textbf{Crossformer}} &\multicolumn{2}{c} {\textbf{TiDe}} & \multicolumn{2}{c}{\textbf{TimesNet}} & \multicolumn{2}{c}{\textbf{DLinear}}  & \multicolumn{2}{c}{\textbf{FEDformer}} \\
\cmidrule(lr){3-4} \cmidrule(lr){5-6} \cmidrule(lr){7-8} \cmidrule(lr){9-10} \cmidrule(lr){11-12} \cmidrule(lr){13-14} \cmidrule(lr){15-16} \cmidrule(lr){17-18} \cmidrule(lr){19-20} \cmidrule(lr){21-22} 
\textbf{Metric} & & \textbf{MSE} & \textbf{MAE} & \textbf{MSE} & \textbf{MAE} & \textbf{MSE} & \textbf{MAE} & \textbf{MSE} & \textbf{MAE} & \textbf{MSE} & \textbf{MAE} & \textbf{MSE} & \textbf{MAE} & \textbf{MSE} & \textbf{MAE} & \textbf{MSE} & \textbf{MAE} & \textbf{MSE} & \textbf{MAE} & \textbf{MSE} & \textbf{MAE}   \\
\midrule
 \multirow{5}{*}{{Weather}} & 96&\textcolor{red}{\textbf{0.161}} & \textcolor{red}{\textbf{0.207}} & 0.165 & 0.210  & \textcolor{blue}{\textbf0.163} & \textcolor{blue}{\textbf0.209} & 0.174 & 0.212 & 0.186 & 0.227 & 0.195 & 0.271 & 0.202 & 0.261 & 0.172 & 0.220  & 0.195 & 0.252 & 0.217 & 0.296 \\
& 192   & \textcolor{red}{\textbf{0.206}} & \textcolor{red}{\textbf{0.246}} & 0.214 & 0.252 & \textcolor{blue}{\textbf0.208} & \textcolor{blue}{\textbf0.250} & 0.221 & 0.254 & 0.234 & 0.265 & 0.209 & 0.277 & 0.242 & 0.298 & 0.219 & 0.261 & 0.237 & 0.282 & 0.276 & 0.336 \\
& 336   & \textcolor{blue}{\textbf0.265} & \textcolor{blue}{\textbf0.291} & 0.274 & 0.297 & \textcolor{red}{\textbf{0.251}} & \textcolor{red}{\textbf{0.278}} & 0.278 & 0.296 & 0.284 & 0.301 & 0.273 & 0.332 & 0.287 & 0.335 & 0.280  & 0.306 & 0.282 & 0.331 & 0.339 & 0.380 \\
& 720   & \textcolor{blue}{\textbf0.344} & \textcolor{red}{\textbf{0.339}} & 0.350  & 0.345 & \textcolor{red}{\textbf{0.339}} & \textcolor{blue}{\textbf0.341} & 0.358 & 0.347 & 0.356 & 0.349 & 0.379 & 0.401 & 0.351 & 0.386 & 0.365 & 0.359 & 0.359 & 0.345 & 0.403 & 0.428 \\
\cmidrule(lr){2-22}  
& \multicolumn{1}{l}{Avg} & \textcolor{blue}{\textbf0.244} & \textcolor{red}{\textbf{0.270}} & 0.251 & 0.276 & \textcolor{red}{\textbf{0.240}}  & \textcolor{blue}{\textbf0.271} & 0.258 & 0.278 & 0.265 & 0.285 & 0.264 & 0.320  & 0.271 & 0.320  & 0.259 & 0.287 & 0.265 & 0.315 & 0.309 & 0.360 \\
\midrule

\multirow{5}{*}{{Electricity}} & 96     & \textcolor{blue}{\textbf0.147} & \textcolor{red}{\textbf{0.233}} & \textcolor{red}{\textbf{0.139}} & \textcolor{blue}{\textbf{0.235}} & 0.153 & 0.247 & 0.148 & 0.240  & 0.190  & 0.296 & 0.219 & 0.314 & 0.237 & 0.329 & 0.168 & 0.272 & 0.210  & 0.305 & 0.169 & 0.273 \\
& 192   & \textcolor{red}{\textbf{0.156}} & \textcolor{red}{\textbf{0.249}} & \textcolor{blue}{\textbf0.159} & 0.255 & 0.166 & 0.256 & 0.162 & \textcolor{blue}{\textbf0.253} & 0.196 & 0.304 & 0.231 & 0.322 & 0.236 & 0.330  & 0.184 & 0.298 & 0.210  & 0.305 & 0.201 & 0.315 \\
& 336   & \textcolor{red}{\textbf{0.174}} & \textcolor{blue}{\textbf0.271} & \textcolor{blue}{\textbf0.176} & 0.272 & 0.185 & 0.277 & 0.178 & \textcolor{red}{\textbf{0.269}} & 0.217 & 0.319 & 0.246 & 0.337 & 0.249 & 0.344 & 0.198 & 0.300   & 0.223 & 0.319 & 0.200   & 0.304 \\
& 720   & {\textcolor{blue}{\textbf0.211}} & \textcolor{red}{\textbf{0.295}} & \textcolor{red}{\textbf{0.204}} & \textcolor{blue}{\textbf0.298} & 0.225 & 0.317 & 0.225 & 0.310  & 0.258 & 0.352 & 0.280  & 0.363 & 0.284 & 0.373 & 0.220  & 0.320  & 0.258 & 0.350  & 0.246 & 0.355 \\
\cmidrule(lr){2-22}  
& \multicolumn{1}{l}{Avg} & \textcolor{blue}{\textbf0.172} & \textcolor{red}{\textbf{0.262}} & \textcolor{red}{\textbf{0.170}}  & \textcolor{blue}{\textbf0.265} & 0.182 & 0.274 & 0.178 & 0.271 & 0.216 & 0.318 & 0.244 & 0.334 & 0.251 & 0.344 & 0.192 & 0.304 & 0.225 & 0.319 & 0.214 & 0.327 \\
\midrule

\multirow{5}{*}{{Exchange}} & 96   & \textcolor{red}{\textbf{0.083}} & \textcolor{red}{\textbf{0.201}} & 0.086 & 0.207 & 0.090  & 0.235 & \textcolor{blue}{\textbf0.086} & 0.206 & 0.088 & \textcolor{blue}{\textbf0.205} & 0.256 & 0.367 & 0.094 & 0.218 & 0.107 & 0.234 & 0.088 & 0.218 & 0.148 & 0.278 \\
& 192   & \textcolor{red}{\textbf{0.173}} & \textcolor{red}{\textbf{0.293}} & 0.182 & 0.304 & 0.187 & 0.343 & \textcolor{blue}{\textbf0.177} & 0.299 & 0.176 & \textcolor{blue}{\textbf0.299} & 0.470  & 0.509 & 0.184 & 0.307 & 0.226 & 0.344 & 0.176 & 0.315 & 0.271 & 0.315 \\
& 336   & \textcolor{blue}{\textbf0.315} & \textcolor{red}{\textbf{0.392}} & 0.332 & 0.418 & 0.353 & 0.473 & 0.331 & 0.417 & \textcolor{red}{\textbf{0.301}} & \textcolor{blue}{\textbf0.397} & 1.268 & 0.883 & 0.349 & 0.431 & 0.367 & 0.448 & 0.313 & 0.427 & 0.460  & 0.427 \\
& 720   & \textcolor{red}{\textbf{0.833}} & \textcolor{red}{\textbf{0.688}} & 0.867 & 0.703 & 0.934 & 0.761 & 0.847 & \textcolor{blue}{\textbf0.691} & 0.901 & 0.714 & 1.767 & 1.068 & 0.852 & 0.698 & 0.964 & 0.746 & \textcolor{blue}{\textbf0.839} & 0.695 & 1.195 & 0.695 \\
\cmidrule(lr){2-22}  
& \multicolumn{1}{l}{Avg} & \textcolor{red}{\textbf{0.351}} & \textcolor{red}{\textbf{0.393}} & 0.367 & 0.408 & 0.391 & 0.453 & \textcolor{blue}{\textbf0.360}  &\textcolor{blue}{\textbf 0.403} & 0.367 & 0.404 & 0.940  & 0.707 & 0.370  & 0.413 & 0.416 & 0.443 & 0.354 & 0.414 & 0.519 & 0.429 \\
\midrule

\multirow{5}{*}{{Solar}} & 96    & \textcolor{blue}{\textbf0.202} & \textcolor{red}{\textbf{0.234}} & 0.205 & 0.244 & \textcolor{red}{\textbf{0.189}} & 0.259 & 0.203 & \textcolor{blue}{\textbf0.237} & 0.234 & 0.286 & 0.310  & 0.331 & 0.312 & 0.399 & 0.250  & 0.292 & 0.290  & 0.378 & 0.242 & 0.342 \\
& 192   & \textcolor{red}{\textbf{0.211}} & \textcolor{red}{\textbf{0.257}} & 0.237 & 0.270  & \textcolor{blue}{\textbf0.222} & 0.283 & 0.233 & \textcolor{blue}{\textbf0.261} & 0.267 & 0.310  & 0.734 & 0.725 & 0.339 & 0.416 & 0.296 & 0.318 & 0.320  & 0.398 & 0.285 & 0.380 \\
& 336   & \textcolor{blue}{\textbf0.246} & \textcolor{blue}{\textbf0.274} & 0.258 & 0.288 & \textcolor{red}{\textbf{0.231}} & 0.292 & 0.248 & \textcolor{red}{\textbf{0.273}} & 0.290  & 0.315 & 0.750  & 0.735 & 0.368 & 0.430  & 0.319 & 0.330  & 0.353 & 0.415 & 0.282 & 0.376 \\
& 720   & \textcolor{blue}{\textbf0.247} & 0.287 & 0.260  & 0.288 & \textcolor{red}{\textbf{0.223}} & \textcolor{blue}{\textbf0.285} & 0.249 & \textcolor{red}{\textbf{0.275}} & 0.289 & 0.317 & 0.769 & 0.765 & 0.370  & 0.425 & 0.338 & 0.337 & 0.356 & 0.413 & 0.357 & 0.427 \\
\cmidrule(lr){2-22}  
& \multicolumn{1}{l}{Avg} & \textcolor{blue}{\textbf0.231} & \textcolor{blue}{\textbf{0.265}} & 0.240  & 0.273 & \textcolor{red}{\textbf{0.216}} & 0.280  & 0.233 & \textcolor{red}{\textbf0.262} & 0.270  & 0.307 & 0.641 & 0.639 & 0.347 & 0.417 & 0.301 & 0.319 & 0.330  & 0.401 & 0.291 & 0.381 \\
\midrule

\multirow{5}{*}{{Traffic}} & 96   & 0.419 & \textcolor{blue}{\textbf0.261} & \textcolor{red}{\textbf{0.382}} & \textcolor{red}{\textbf{0.261}} & 0.462 & 0.285 & \textcolor{blue}{\textbf0.395} & 0.271 & 0.526 & 0.347 & 0.644 & 0.429 & 0.805 & 0.493 & 0.593 & 0.321 & 0.650  & 0.396 & 0.587 & 0.366 \\
& 192   & 0.428 & 0.282 & \textcolor{red}{\textbf{0.396}} & \textcolor{red}{\textbf{0.267}} & 0.473 & 0.296 & \textcolor{blue}{\textbf0.417} & \textcolor{blue}{\textbf0.276} & 0.522 & 0.332 & 0.665 & 0.431 & 0.756 & 0.474 & 0.617 & 0.336 & 0.598 & 0.370 & 0.604 & 0.373 \\
& 336   & 0.466 & \textcolor{blue}{\textbf0.294} & \textcolor{red}{\textbf{0.417}} & \textcolor{red}{\textbf{0.276}} & 0.498 & 0.296 & \textcolor{blue}{\textbf0.433} & 0.298 & 0.517 & 0.334 & 0.674 & 0.420  & 0.762 & 0.477 & 0.629 & 0.336 & 0.605 & 0.373 & 0.621 & 0.383 \\
& 720   & 0.482 & \textcolor{blue}{\textbf0.301} & \textcolor{red}{\textbf{0.460}} & \textcolor{red}{\textbf{0.300}} & 0.506 & 0.313 & \textcolor{blue}{\textbf0.467} & 0.302 & 0.552 & 0.352 & 0.683 & 0.424 & 0.719 & 0.449 & 0.640  & 0.350  & 0.645 & 0.394 & 0.626 & 0.382 \\
\cmidrule(lr){2-22}  
& \multicolumn{1}{l}{Avg} & 0.478 & \textcolor{blue}{\textbf0.286} & \textcolor{red}{\textbf{0.414}} & \textcolor{red}{\textbf{0.276}} & 0.484 & 0.298 & \textcolor{blue}{\textbf0.428} & 0.287 & 0.529 & 0.341 & 0.667 & 0.426 & 0.760  & 0.473 & 0.620  & 0.336 & 0.625 & 0.383 & 0.610  & 0.376 \\
\bottomrule
\end{tabular}
}
\label{tab:main_results}
\end{sidewaystable*}

\begin{sidewaystable*}[p]
\centering
\renewcommand{\arraystretch}{1.4}
\setlength{\tabcolsep}{1.1pt} 
\addtocounter{table}{-1}
\caption{(continued)}
\resizebox{\linewidth}{!}{%
 \begin{tabular}{l l *{20}{w{r}{3em}}} 
\toprule
\multicolumn{2}{l}{\textbf{Models}} & \multicolumn{2}{c}{\textbf{ASGMamba}} & \multicolumn{2}{c}{\textbf{S-Mamba}} & \multicolumn{2}{c}{\textbf{TimeMixer}} & \multicolumn{2}{c}{\textbf{iTransformer}} & \multicolumn{2}{c}{\textbf{PatchTST}} & \multicolumn{2}{c}{\textbf{Crossformer}} &\multicolumn{2}{c} {\textbf{TiDe}} & \multicolumn{2}{c}{\textbf{TimesNet}} & \multicolumn{2}{c}{\textbf{DLinear}}  & \multicolumn{2}{c}{\textbf{FEDformer}} \\

\cmidrule(lr){3-4} \cmidrule(lr){5-6} \cmidrule(lr){7-8} \cmidrule(lr){9-10} \cmidrule(lr){11-12} \cmidrule(lr){13-14} \cmidrule(lr){15-16} \cmidrule(lr){17-18} \cmidrule(lr){19-20} \cmidrule(lr){21-22} 
\textbf{Metric} & & \textbf{MSE} & \textbf{MAE} & \textbf{MSE} & \textbf{MAE} & \textbf{MSE} & \textbf{MAE} & \textbf{MSE} & \textbf{MAE} & \textbf{MSE} & \textbf{MAE} & \textbf{MSE} & \textbf{MAE} & \textbf{MSE} & \textbf{MAE} & \textbf{MSE} & \textbf{MAE} & \textbf{MSE} & \textbf{MAE}  & \textbf{MSE} & \textbf{MAE}  \\
\midrule
\multirow{5}{*}{{ETTh1}} & 96    & \textcolor{red}{\textbf{0.369}} & \textcolor{red}{\textbf{0.391}} & 0.386 & 0.405 & \textcolor{blue}{\textbf0.375} & \textcolor{blue}{\textbf0.400} & 0.386 & 0.405 & 0.460  & 0.447 & 0.423 & 0.448 & 0.479 & 0.464 & 0.384 & 0.402 & 0.407 & 0.412 & 0.395 & 0.424 \\
& 192   & \textcolor{red}{\textbf{0.426}} & \textcolor{red}{\textbf{0.419}} & 0.443 & 0.437 & \textcolor{blue}{\textbf0.429} & {\textcolor{blue}{\textbf0.421}} & 0.441 & 0.512 & 0.477 & 0.429 & 0.471 & 0.474 & 0.525 & 0.492 & 0.436 & 0.446 & 0.441 & 0.411 & 0.469 & 0.470 \\
& 336   & \textcolor{blue}{\textbf0.478} & \textcolor{red}{\textbf{0.444}} & 0.489 & 0.468 & \textcolor{red}{\textbf{0.458}} & \textcolor{blue}{\textbf0.458} & 0.487 & 0.458 & 0.546 & 0.496 & 0.496 & 0.470  & 0.565 & 0.515 & 0.491 & 0.491 & 0.469 & 0.489 & 0.547 & 0.495 \\
& 720   & \textcolor{red}{\textbf{0.487}} & \textcolor{red}{\textbf{0.473}} & 0.502 & 0.489 & \textcolor{blue}{\textbf0.498} & \textcolor{blue}{\textbf0.482} & 0.503 & 0.491 & 0.544 & 0.517 & 0.653 & 0.621 & 0.594 & 0.558 & 0.521 & 0.500   & 0.513 & 0.510  & 0.598 & 0.544 \\
\cmidrule(lr){2-22}  
& \multicolumn{1}{l}{Avg} & \textcolor{red}{\textbf{0.440}} & \textcolor{red}{\textbf{0.432}} & 0.455 & 0.450  & \textcolor{blue}{\textbf0.447} & \textcolor{blue}{\textbf0.440} & 0.454 & 0.447 & 0.516 & 0.484 & 0.529 & 0.522 & 0.541 & 0.507 & 0.458 & 0.450  & 0.461 & 0.457 & 0.498 & 0.484 \\
\midrule

\multirow{5}{*}{{ETTh2}} & 96   & \textcolor{red}{\textbf{0.283}} & \textcolor{red}{\textbf{0.335}} & 0.296 & 0.348 & 0.289 & 0.342 & \textcolor{blue}{\textbf0.286} & \textcolor{blue}{\textbf0.338} & 0.745 & 0.584 & 0.745 & 0.584 & 0.400   & 0.440  & 0.340  & 0.374 & 0.340  & 0.394 & 0.358 & 0.397 \\
& 192   & \textcolor{red}{\textbf{0.367}} & \textcolor{red}{\textbf{0.391}} & 0.376 & \textcolor{blue}{\textbf0.396} & \textcolor{blue}{\textbf0.372} & 0.397 & 0.380  & 0.400   & 0.793 & 0.585 & 0.877 & 0.656 & 0.528 & 0.509 & 0.402 & 0.452 & 0.419 & 0.479 & 0.414 & 0.439 \\
& 336   & \textcolor{blue}{\textbf0.411} & \textcolor{red}{\textbf{0.412}} & 0.424 & 0.431 & \textcolor{red}{\textbf{0.386}} & \textcolor{blue}{\textbf0.414} & 0.428 & 0.432 & 0.927 & 0.643 & 1.043 & 0.731 & 0.643 & 0.571 & 0.452 & 0.482 & 0.591 & 0.541 & 0.496 & 0.487 \\
& 720   & \textcolor{red}{\textbf{0.409}} & \textcolor{red}{\textbf{0.431}} & 0.426 & 0.444 & \textcolor{blue}{\textbf0.412} & \textcolor{blue}{\textbf0.434} & 0.427 & 0.445 & 1.043 & 0.636 & 1.104 & 0.763 & 0.874 & 0.679 & 0.462 & 0.468 & 0.661 & 0.661 & 0.463 & 0.474 \\
\cmidrule(lr){2-22}  
& \multicolumn{1}{l}{Avg} & \textcolor{red}{\textbf{0.367}} & \textcolor{red}{\textbf{0.392}} & 0.381 & 0.405 & \textcolor{blue}{\textbf0.364} & \textcolor{blue}{\textbf0.395} & 0.383 & 0.407 & 0.878 & 0.612 & 0.841 & 0.642 & 0.611 & 0.550  & 0.414 & 0.427 & 0.563 & 0.519 & 0.437 & 0.449 \\
\midrule
\multirow{5}{*}{{ETTm1}} & 96     & \textcolor{red}{\textbf{0.315}} & \textcolor{red}{\textbf{0.353}} & 0.333 & 0.368 & \textcolor{blue}{\textbf0.320} & \textcolor{blue}{\textbf0.357} & 0.334 & 0.368 & 0.352 & 0.374 & 0.404 & 0.426 & 0.364 & 0.387 & 0.338 & 0.375 & 0.346 & 0.374 & 0.379 & 0.419 \\
& 192   & \textcolor{red}{\textbf{0.357}} & \textcolor{red}{\textbf{0.385}} & 0.376 & \textcolor{blue}{\textbf0.390} & \textcolor{blue}{\textbf0.361} & 0.393 & 0.404 & 0.393 & 0.387 & 0.404 & 0.450  & 0.451 & 0.398 & 0.404 & 0.374 & 0.387 & 0.381 & 0.391 & 0.389 & 0.387 \\
& 336   & \textcolor{red}{\textbf{0.385}} & \textcolor{red}{\textbf{0.401}} & 0.408 & 0.413 & \textcolor{blue}{\textbf0.390} & \textcolor{blue}{\textbf0.404} & 0.426 & 0.420  & 0.421 & 0.414 & 0.532 & 0.515 & 0.428 & 0.425 & 0.410  & 0.411 & 0.415 & 0.415 & 0.445 & 0.459 \\
& 720   & \textcolor{red}{\textbf{0.449}} & \textcolor{red}{\textbf{0.436}} & 0.475 & 0.448 & \textcolor{blue}{\textbf0.454} & \textcolor{blue}{\textbf0.441} & 0.491 & 0.459 & 0.462 & 0.449 & 0.666 & 0.589 & 0.487 & 0.461 & 0.478 & 0.450  & 0.473 & 0.451 & 0.543 & 0.490 \\
\cmidrule(lr){2-22}        
& \multicolumn{1}{l}{Avg} & \textcolor{red}{\textbf{0.376}} & \textcolor{red}{\textbf{0.393}} & 0.398 & 0.405 & \textcolor{blue}{\textbf0.381} & \textcolor{blue}{\textbf0.398} & 0.407 & 0.410  & 0.406 & 0.407 & 0.513 & 0.495 & 0.419 & 0.419 & 0.400   & 0.406 & 0.404 & 0.408 & 0.448 & 0.452 \\

\midrule
\multirow{5}{*}{{ETTm2}} & 96    & \textcolor{red}{\textbf{0.171}} & \textcolor{red}{\textbf{0.252}} & 0.179 & 0.263 & \textcolor{blue}{\textbf0.175} & \textcolor{blue}{\textbf0.258} & 0.180  & 0.264 & 0.183 & 0.270  & 0.287 & 0.366 & 0.207 & 0.305 & 0.187 & 0.267 & 0.193 & 0.286 & 0.203 & 0.287 \\
& 192   & \textcolor{red}{\textbf{0.233}} & \textcolor{red}{\textbf{0.293}} & 0.250  & 0.309 & \textcolor{blue}{\textbf0.237} & \textcolor{blue}{\textbf0.299} & 0.250  & 0.309 & 0.255 & 0.314 & 0.414 & 0.492 & 0.290  & 0.364 & 0.249 & 0.309 & 0.284 & 0.361 & 0.269 & 0.328 \\
& 336   & \textcolor{red}{\textbf{0.294}} & \textcolor{red}{\textbf{0.332}} & 0.312 & 0.349 & \textcolor{blue}{\textbf0.298} & \textcolor{blue}{\textbf0.340} & 0.311 & 0.348 & 0.309 & 0.347 & 0.597 & 0.542 & 0.377 & 0.422 & 0.321 & 0.331 & 0.382 & 0.429 & 0.325 & 0.366 \\
& 720   & \textcolor{blue}{\textbf0.401} & \textcolor{blue}{\textbf0.402} & 0.411 & 0.406 & \textcolor{red}{\textbf{0.391}} & \textcolor{red}{\textbf{0.392}} & 0.407 & 0.407 & 0.412 & 0.404 & 1.730  & 1.042 & 0.558 & 0.524 & 0.408 & 0.403 & 0.558 & 0.525 & 0.421 & 0.415 \\
\cmidrule(lr){2-22}  
& \multicolumn{1}{l}{Avg} & \textcolor{red}{\textbf{0.274}} & \textcolor{red}{\textbf{0.319}} & 0.288 & 0.332 & \textcolor{blue}{\textbf0.275} & \textcolor{blue}{\textbf0.323} & 0.288 & 0.332 & 0.290  & 0.334 & 0.757 & 0.610  & 0.358 & 0.404 & 0.291 & 0.333 & 0.354 & 0.402 & 0.305 & 0.349 \\
\midrule
\multicolumn{2}{l}{1st Count} & \textbf{21}    & \textbf{27}    & 6     & 4     & 8     & 2     & 0     & 3     & 1     & 0     & 0     & 0     & 0     & 0     & 0     & 0     & 0     & 0     & 0     & 0 \\
\bottomrule
\end{tabular}
}
\label{tab:full_long_term_forecasting}
\end{sidewaystable*}

\subsection{Main Results}

Table \ref{tab:full_long_term_forecasting} presents a comprehensive quantitative evaluation of ASGMamba against ten state-of-the-art baselines across nine diverse real-world datasets. The experiments cover four distinct prediction horizons $T \in \{96, 192, 336, 720\}$, providing a rigorous assessment of Long-term Time Series Forecasting (LTSF) capabilities. As evidenced by the aggregated results, ASGMamba achieves the lowest average Mean Squared Error (MSE) on five out of the nine datasets and the lowest average Mean Absolute Error (MAE) on seven out of the nine datasets. This positions ASGMamba as effective in linear-complexity forecasting.

\textbf{Superiority in Volatile Environments.} 
ASGMamba outperforms advanced Transformer-based architectures on datasets with low Signal-to-Noise Ratios (SNR). For instance, on the Exchange dataset, which features severe structural breaks, ASGMamba records an average MSE of 0.351 outperforming iTransformer (0.360) and PatchTST (0.367). Similarly, on the Ettm1 dataset, ASGMamba surpasses both the vanilla S-Mamba (0.398) and TimeMixer (0.381) with an MSE of 0.376. These results validate our core hypothesis: in scenarios laden with stochastic perturbations, the proposed Adaptive Spectral Gating mechanism effectively filters high-frequency noise, allowing the model to capture robust long-term trends that standard attention mechanisms often miss due to overfitting.

\textbf{Limitations on High-Dimensional Data.} 
However, the performance landscape shifts on high-dimensional datasets with massive channel counts, such as Traffic (862 variates) and Electricity (321 variates). While ASGMamba remains competitive against PatchTST, it yields suboptimal results compared to iTransformer and the vanilla S-Mamba. On Traffic, ASGMamba achieves an average MSE of 0.478, trailing behind S-Mamba (0.414). We attribute this to the aggressive nature of our gating mechanism. Traffic data is characterized by high-entropy, non-stationary fluctuations that are physically meaningful (e.g., sudden congestion) but spectrally resemble noise. The gating mechanism may inadvertently suppress these high-frequency signals, leading to information loss. Furthermore, the vanilla S-Mamba utilizes a bidirectional scanning mechanism across variates, which naturally excels at capturing dense spatial correlations in high-dimensional settings, whereas our explicit Node Embedding strategy may face capacity bottlenecks when scaling to nearly a thousand sensors.

\textbf{Mechanism Validation and Stability.} 
Despite the trade-off in high-dimensional settings, ASGMamba demonstrates superior stability on the ETT family datasets compared to the baseline S-Mamba. On ETTh1 and ETTm1, our model improves upon S-Mamba by approximately 3.1\% and 1.3\% respectively. This confirms that for datasets where variables operate with distinct physical semantics but moderate dimensionality, the combination of spectral gating and semantic node embeddings provides a clear advantage over raw state space modeling. Additionally, ASGMamba exhibits exceptional robustness at the ultra-long horizon ($T=720$). For example, on ETTm2, it achieves an MSE of 0.401, matching the performance of specialized decomposition models like TimeMixer and significantly outperforming older baselines like FEDformer (0.421). In summary, ASGMamba offers a specialized trade-off: it sacrifices some granularity on massive-channel datasets to achieve strong noise robustness and trend capture capability in volatile, low-SNR environments.

\begin{figure}[htbp]
    \centering %
    \begin{minipage}[b]{0.48\columnwidth}
        \centering
        \includegraphics[width=\linewidth]{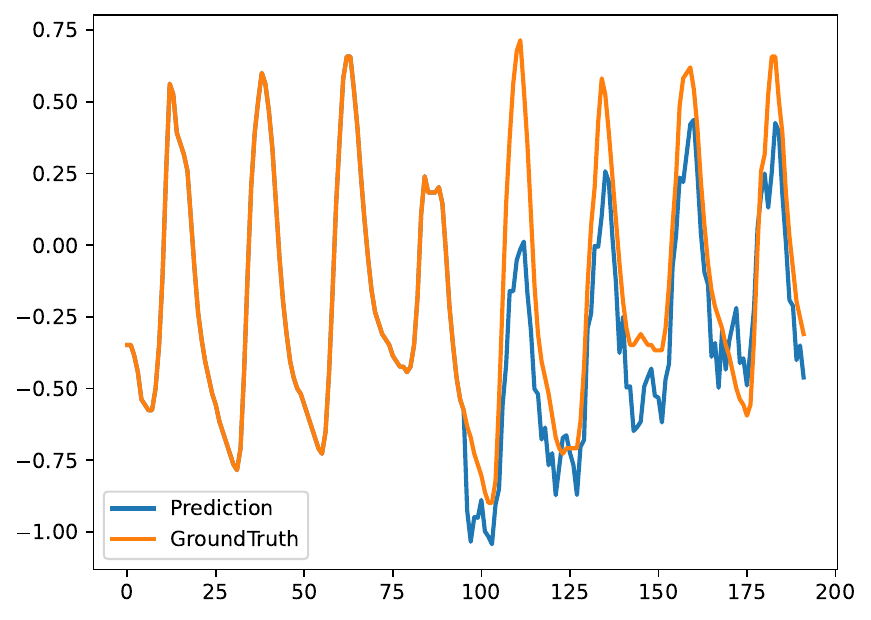}
        \centerline{(a) Etth2-96-ASGMamba}
    \end{minipage}%
    \hfill
   \begin{minipage}[b]{0.48\columnwidth}
    \centering
    \includegraphics[width=\linewidth]{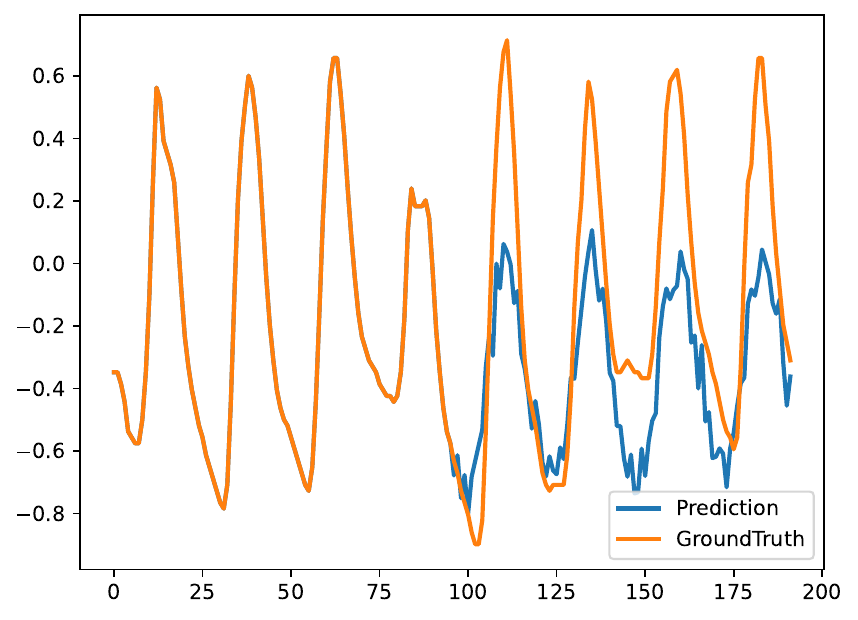}
    \centerline{(d) Etth2-96-SMamba}
    \end{minipage}
\vspace{0.3cm} %
    \begin{minipage}[b]{0.48\columnwidth}
        \centering
        \includegraphics[width=\linewidth]{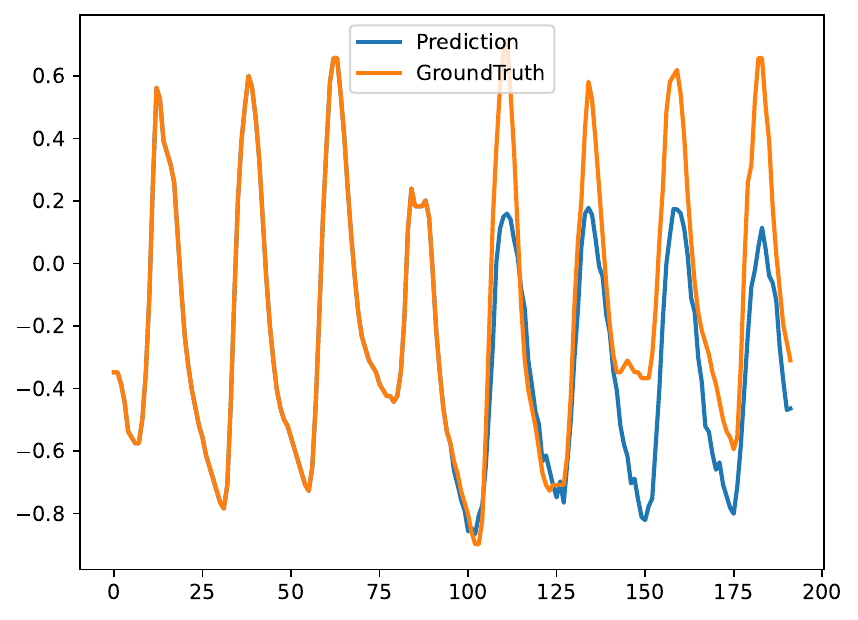}
        \centerline{(c) Etth2-96-TimeMixer}
    \end{minipage}%
    \hfill
     \begin{minipage}[b]{0.48\columnwidth}
        \centering
        \includegraphics[width=\linewidth]{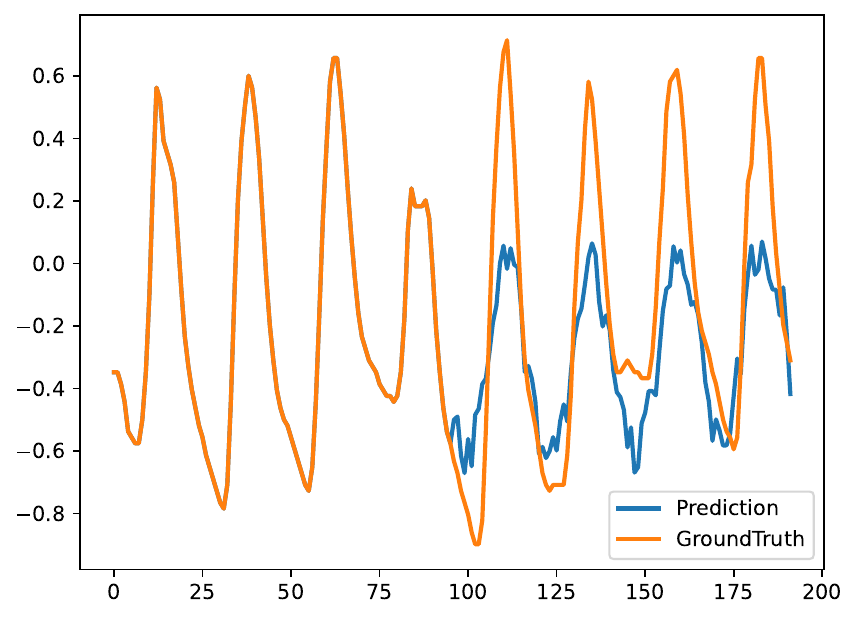}
        \centerline{(b) Etth2-96-iTransformer}
    \end{minipage}

\vspace{0.3cm} 

    \begin{minipage}[b]{0.48\columnwidth}
        \centering
        \includegraphics[width=\linewidth]{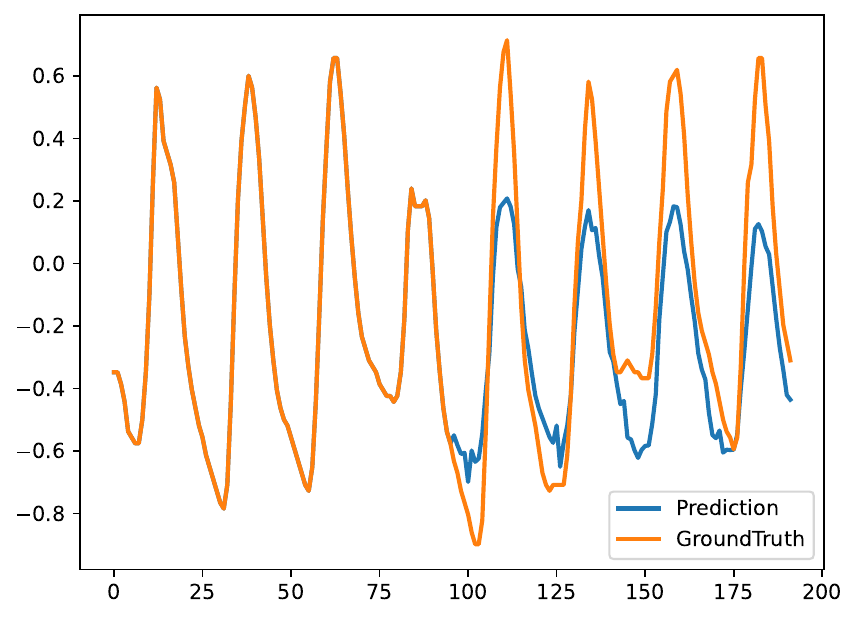}
        \centerline{(e) Etth2-96-CrossFormer}
    \end{minipage}%
    \hfill
    \begin{minipage}[b]{0.48\columnwidth}
        \centering
        \includegraphics[width=\linewidth]{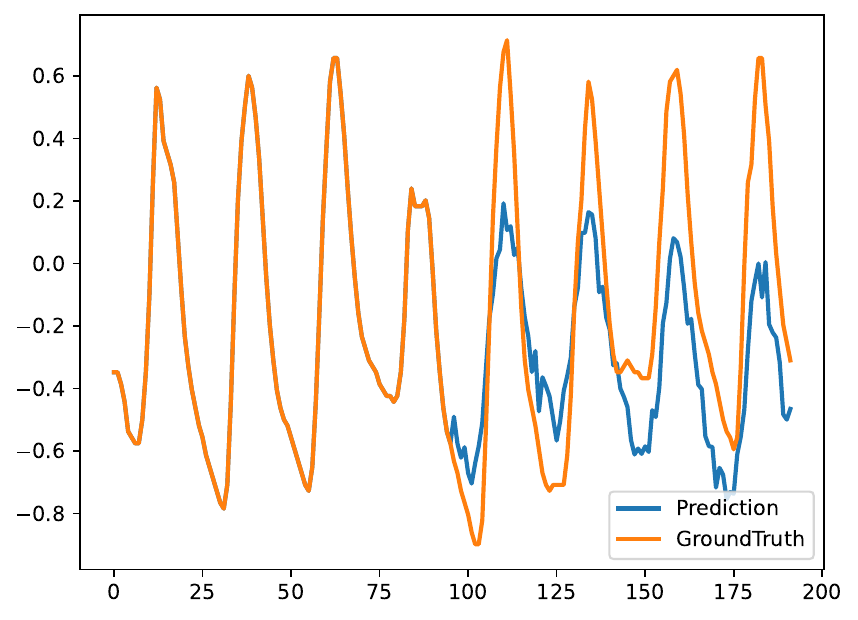}
        \centerline{(f) Etth2-96-PatchTST}
    \end{minipage}
    \vspace{0.3cm} 

    \begin{minipage}[b]{0.48\columnwidth}
        \centering
        \includegraphics[width=\linewidth]{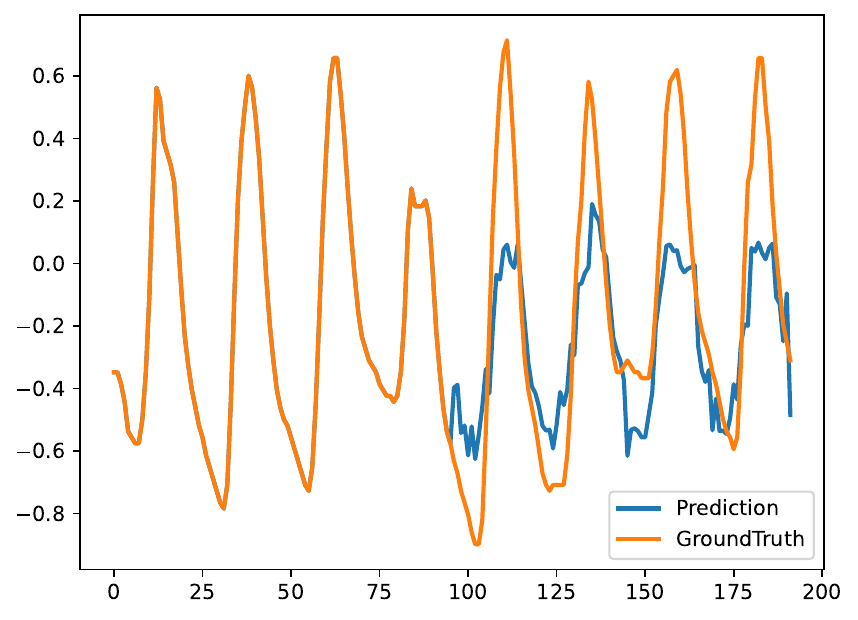}
        \centerline{(g) Etth2-96-TimesNet}
    \end{minipage}%
    \hfill
    \begin{minipage}[b]{0.48\columnwidth}
        \centering
        \includegraphics[width=\linewidth]{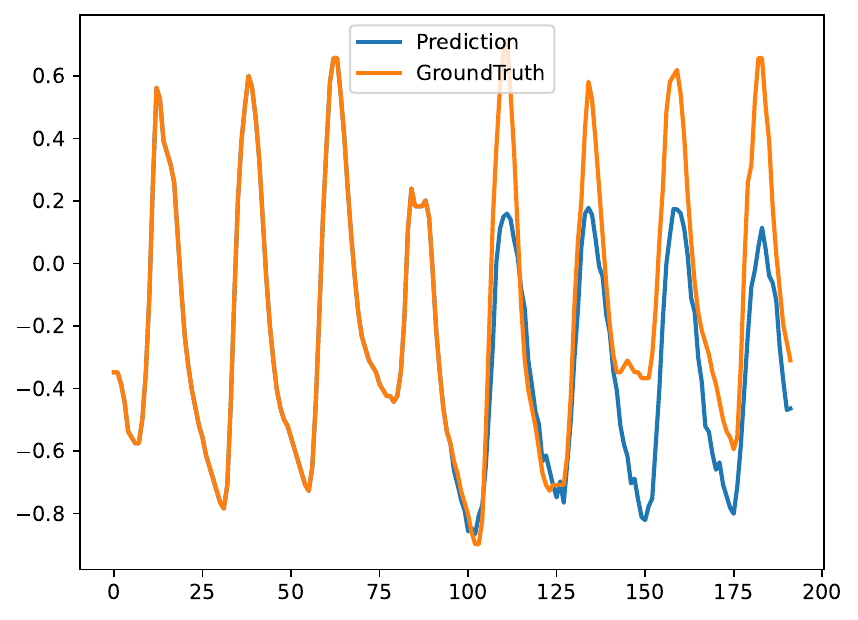}
        \centerline{(h) Etth2-96-Dlinear}
    \end{minipage}
  \caption{Visual comparison of forecasting performance on the Etth2 dataset with a horizon of $T=96$.}
    \label{fig:etth2_96}
\end{figure}

\subsection{Visualizing and Analyzing}

To intuitively evaluate the temporal modeling capability of ASGMamba, we visualize the forecasting results on the ETTh2 dataset with a prediction horizon of $T=96$. Figure \ref{fig:etth2_96} presents a qualitative comparison between ASGMamba and seven representative baselines, including S-Mamba, iTransformer, and PatchTST. The ETTh2 dataset is characterized by strong cyclicity mixed with irregular amplitude shifts and local fluctuations, posing a dual challenge: models must capture the dominant seasonal trend without overfitting to high-frequency stochastic noise or lagging in phase during rapid transitions.

As illustrated in Fig. \ref{fig:etth2_96}(a), ASGMamba demonstrates superior alignment with the Ground Truth curve (orange). Specifically, it accurately reconstructs the sharp peaks and troughs of the load cycle with minimal phase shift. In contrast, the vanilla S-Mamba (Fig. \ref{fig:etth2_96}(d)) and linear baselines like DLinear (Fig. \ref{fig:etth2_96}(h)) exhibit noticeable amplitude decay and "smoothing effects" as the prediction steps increase, failing to capture the full intensity of peak loads. While advanced Transformers like PatchTST (Fig. \ref{fig:etth2_96}(f)) and iTransformer (Fig. \ref{fig:etth2_96}(b)) successfully model the general trend, they show slight deviations in local details and instability at the boundaries of the prediction window compared to our method.

The visual superiority supports the theoretical benefits of the proposed Adaptive Spectral Gating mechanism. By analyzing local spectral energy, ASGMamba effectively filters out unnecessary high-frequency noise before it enters the state space. This denoising process enables the selective scan mechanism to focus its limited capacity on preserving strong periodic patterns, rather than fitting noise. As a result, ASGMamba produces forecasts that are not only statistically accurate but also consistent with the underlying dynamic system, avoiding the spurious fluctuations commonly seen in time-domain models.

\subsection{Ablation Studies}
\label{sec:ablation}

To disentangle the contribution of individual components within ASGMamba and validate our design choices, we conducted a comprehensive ablation analysis. We selected ETTh1 and Weather as representative benchmarks due to their contrasting statistical properties: ETTh1 is characterized by strong seasonality corrupted by noise, whereas Weather exhibits complex, non-linear dynamics with high volatility.
We evaluated the full model against four variants, systematically isolating the Adaptive Spectral Gating, Multi-Scale Fusion, Overlapping Patching, and Gating Strategies. The quantitative results are presented in Table~\ref{tab:ablation}.

\begin{sidewaystable*}[p]
\caption{Ablation results for different ASGMamba modules on ETTh1 and Weather. The table displays MSE and MAE values, as well as the relative performance deterioration (\%) for each ablated part.}
\label{tab:ablation}
\centering
\resizebox{1.0\textwidth}{!}{
\setlength{\tabcolsep}{0.9em} 
\begin{tabular}{llcccccccccc}
\toprule
\multicolumn{2}{c}{\textbf{Model Variant}} & \multicolumn{2}{c}{\textbf{Ours}} & \multicolumn{2}{c}{\textbf{w/o  Spectral Gating}} & \multicolumn{2}{c}{\textbf{w/o Multi-Scale Fusion}} & \multicolumn{2}{c}{\textbf{w/o Patching}} & \multicolumn{2}{c}{\textbf{w/o Gating}}  \\
\cmidrule(lr){3-4} \cmidrule(lr){5-6} \cmidrule(lr){7-8} \cmidrule(lr){9-10} \cmidrule(lr){11-12}
\textbf{Metric} & & MSE & MAE & MSE & MAE & MSE & MAE & MSE & MAE & MSE & MAE \\
\midrule
\multirow{4}{*}{Weather} 
& 96  & \textbf{0.161} & \textbf{0.207} & 0.169 & 0.221 & 0.169 & 0.218 & 0.166 & 0.213 & 0.165 & 0.215  \\
& 192 & \textbf{0.206} & \textbf{0.246} & 0.212 & 0.252 & 0.216 & 0.254 & 0.212 & 0.251 & 0.209 & 0.252  \\
& 336 & \textbf{0.265} & \textbf{0.291} & 0.268 & 0.301 & 0.275 & 0.305 & 0.272 & 0.301 & 0.273 & 0.298  \\
& 720 & \textbf{0.344} & \textbf{0.339} & 0.352 & 0.343 & 0.356 & 0.353 & 0.346 & 0.350 & 0.348 & 0.346  \\
\cmidrule(lr){2-12}
 Degradation & & - & - & 2.83\% & 3.45\% & 4.27 \% & 4.38 \%  & 2.31\%  & 2.90 \% & 2.03 \% & 2.69 \%   \\
\midrule
\multirow{4}{*}{Etth1} 
& 96  & \textbf{0.369} & \textbf{0.391} & 0.382 & 0.402 & 0.392 & 0.406 & 0.381 & 0.393 & 0.381 & 0.398   \\
& 192 & \textbf{0.426} & \textbf{0.419} & 0.438 & 0.432 & 0.439 & 0.442 & 0.436 & 0.435 & 0.438 & 0.426   \\
& 336 & \textbf{0.478} & \textbf{0.444} & 0.499 & 0.456 & 0.496 & 0.468 & 0.488 & 0.453 & 0.489 & 0.456   \\
& 720 & \textbf{0.487} & \textbf{0.473} & 0.502 & 0.495 & 0.516 & 0.496 & 0.493 & 0.482 & 0.496 & 0.486  \\
\cmidrule(lr){2-12}
Degradation&  & - & - & {3.45 \%} & 3.32 \% & 4.75 \% & 4.90\% & 2.23\% & 2.06\% & 2.56 \% & 2.16\% \\
\bottomrule
\end{tabular}
}
\end{sidewaystable*}

\subsubsection{Impact of Adaptive Spectral Gating}
The Spectral Gating structurally distinguish robust periodic patterns from high-frequency stochastic noise. In the w/o Spectral Gating variant, we bypassed this mechanism (setting gating weights $\mathbf{G} \equiv \mathbf{1}$), effectively reverting the backbone to a standard Mamba block.

As shown in Table~\ref{tab:ablation}, removing FAR brings about certain performance drop. Specifically, we find that MSE increases about 3.45\% on ETTh1 and 2.83\% on Weather on average. Such uniform drop verifies our hypothesis on standard SSMs limitation: without spectral guidance, the selective scan mechanism treats input tokens as an entire pool. For datasets containing abundant stochastic perturbation (e.g., Weather), to fit high-frequency noise, the model misuses precious state capacity. By introducing FAR, ASGMamba plays the role of an adaptive spectral filter that suppresses noise-dominated components in the frequency domain from invading state space before they pollute unseen future horizons and improve generalization.

\subsubsection{Efficacy of Adaptive Multi-Scale Fusion}

Semantic temporal patterns naturally appear at different granularities, which present a dilemma between local resolution and global context. ASGMamba alleviates this by designing a three-branch architecture ($P \in \{8, 16, 32\}$). w/o Multi-Scale Fusion is forced to rely on only one scale ($P=16$).
The accuracy drop is significant, which is even larger than that caused by other modules. That is, we can observe an approximate 4.75\% MSE increase on ETTh1 and {4.27\%} on Weather. We attribute this to the fixed receptive field of single-scale-patches. Specifically, small patches ($P=8$) capture fine-grained jitter while they lack global context for long-term dependency modeling; while large patches ($P=32$) capture global trend but suffer from spectral smoothing, leading to loss of high-frequency information. ASGMamba adapts to dynamically aggregating these two views. The model can focus on intrinsic periodicity and volatility of different variables and assign different weights.

\subsubsection{Significance of Overlapping Patching}
Conventional non-overlapping patching ($\text{Stride} = \text{Patch}$) introduces artificial "boundary artifacts" disrupting the semantic continuity of local temporal patterns. We validated the proposed Overlapping Patching strategy ($\text{Stride} = \text{Patch}/2$) against the non-overlapping baseline.

While the performance gain (approx. 2.23\% on ETTh1) is more modest compared to the Multi-Scale architecture, the consistent improvement across datasets underscores its necessity. By enforcing a 50\% overlap, we ensure that critical transition points are encoded in multiple patches with varying relative positions. This redundancy provides the Mamba encoder with enriched contextual views, mitigating information loss at patch boundaries and fostering smoother latent state transitions.

\subsubsection{Sensitivity to Gating Strategies}
We assess whether gains stem from the spectral domain by comparing variants. Specifically, we benchmark our spectral gating against a non-gated baseline (Identity) and a variant devoid of frequency-aware capabilities.

Empirically, the w/o Gating variant yielded a smaller degradation (approx. 2.0\% on Weather) compared to the removal of the Adaptive Spectral Gating logic (2.83\%), while the full ASGMamba achieves the lowest MSE (e.g., {0.369} on ETTh1-96). This disparity highlights the fundamental advantage of spectral analysis in signal separation. In the time domain (or simple identity mapping), signal and noise are intricately entangled, making it difficult to distinguish a valid anomaly from a random spike. In contrast, stochastic noise typically exhibits a distinct "broadband" energy signature in the frequency domain. This spectral distinctiveness enables the Gating Mechanism to identify and suppress noise components more robustly than standard gating approaches, justifying the coupling of FFT with the gating mechanism.

\subsection{Hyper-parameter Sensitivity Analysis}
\label{subsec:sensitivity}

To verify the robustness of ASGMamba and identify the optimal configuration for forecasting tasks, we conducted a comprehensive sensitivity analysis on two critical hyper-parameters: SSM State Dimension ($N$) and the Spectral Bands ($K_{freq}$). All experiments were performed on the Weather dataset with a prediction horizon of $T=96$.

\begin{figure}[htbp]   
    \centering
    \begin{minipage}{0.48\linewidth}  
        \centering
        \  \includegraphics[width=0.95\columnwidth]{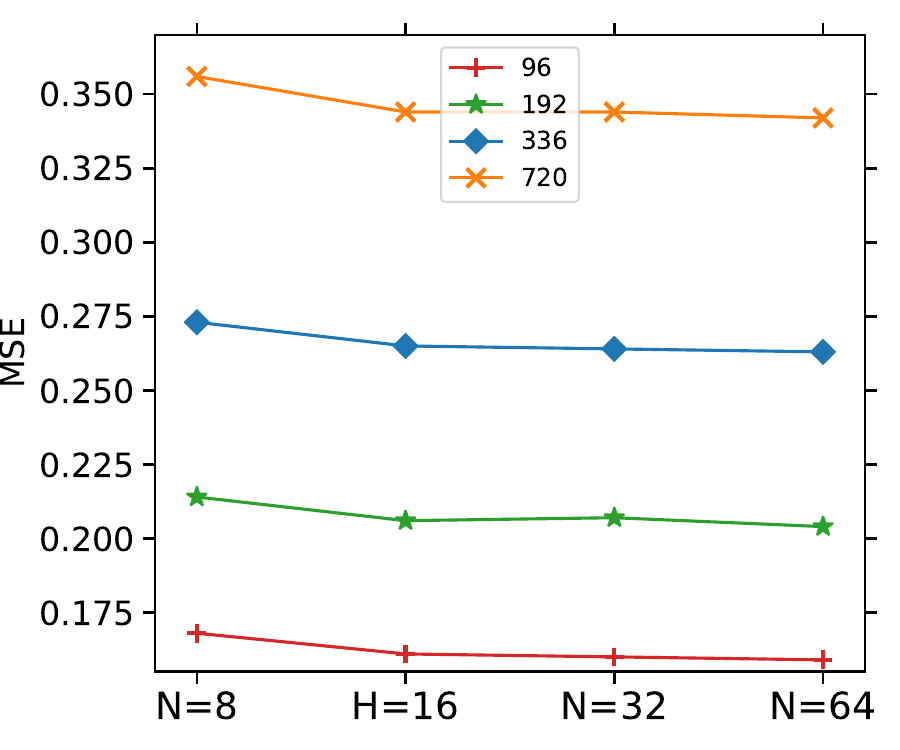}
        \centerline{(a) Impact of SSM state dimension $N$.}
    \end{minipage}%
    \hfill
    \begin{minipage}{0.48\linewidth}
        \centering
        \includegraphics[width=0.95\columnwidth]{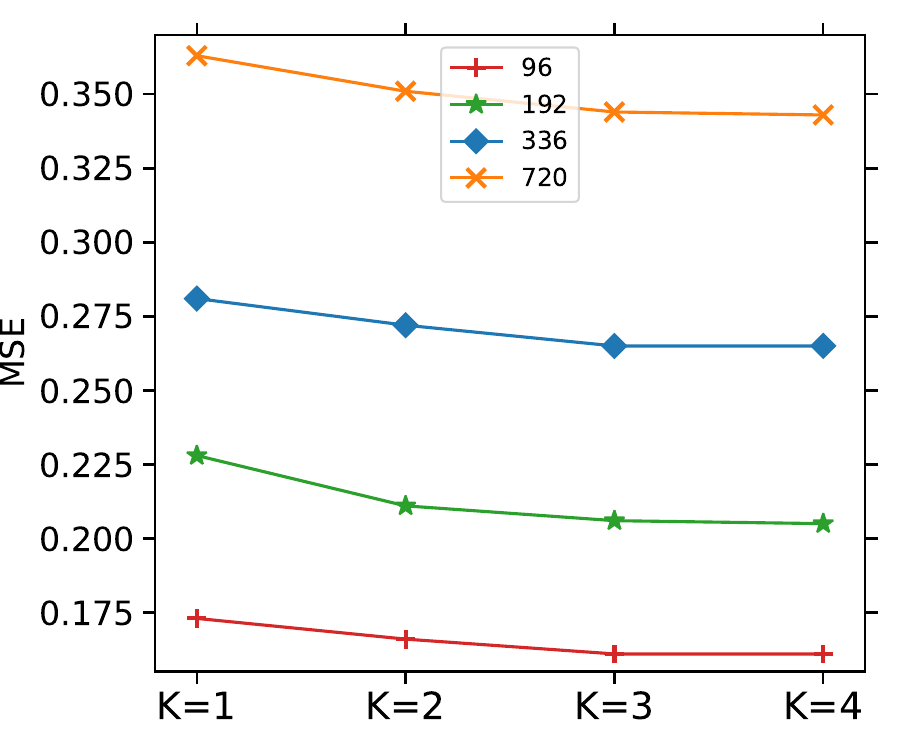} 
        \centerline{(b) Impact of spectral band granularity $K_{freq}$.}
    \end{minipage}
    \caption{Hyperparameter sensitivity analysis on the Weather dataset.}
    \label{fig:hyper_params}
\end{figure}

\noindent\textbf{Impact of SSM State Dimension} 
The state dimension $N$ controls the memory capacity of the selective scan mechanism. We tested $N \in \{8, 16, 32, 64\}$. The results in Fig.~\ref{fig:hyper_params}(a) demonstrate that ASGMamba is relatively robust to $N$. While $N=8$ underperforms due to constrained information bottlenecks, $N=16$ achieves competitive results comparable to larger settings ($N=32, 64$). This suggests that the Adaptive Spectral Gating efficiently filters noise, allowing the SSM to store compact, robust dynamics without requiring an excessively large state space.

\noindent\textbf{Impact of Spectral Band Granularity}

The number of spectral bands into which the spectrum is divided is equal to the resolution of the noise filter. Therefore, we also tried to divide the spectrum into $K_{freq} \in \{1, 2, 3, 4\}$ bands. As illustrated in Fig.~\ref{fig:hyper_params}(b), the case of $K_{freq}=1$ (Global Energy) is suboptimal because the gating mechanism cannot distinguish high energy informative signals from high energy artifacts. The accuracy improves until $K_{freq}=3$. Physically, this division matches the common sense of time series decomposition. Indeed, the spectrum is properly separated into three parts: Trend (Low), Periodicity (Mid), and Noise (High). When $K_{freq}=4$, there is an additional cost without any improvement in performance, which confirms that $K_{freq}=3$  is the optimal setting.
\begin{figure}[htbp]
    \centering
    \includegraphics[width=0.95\columnwidth]{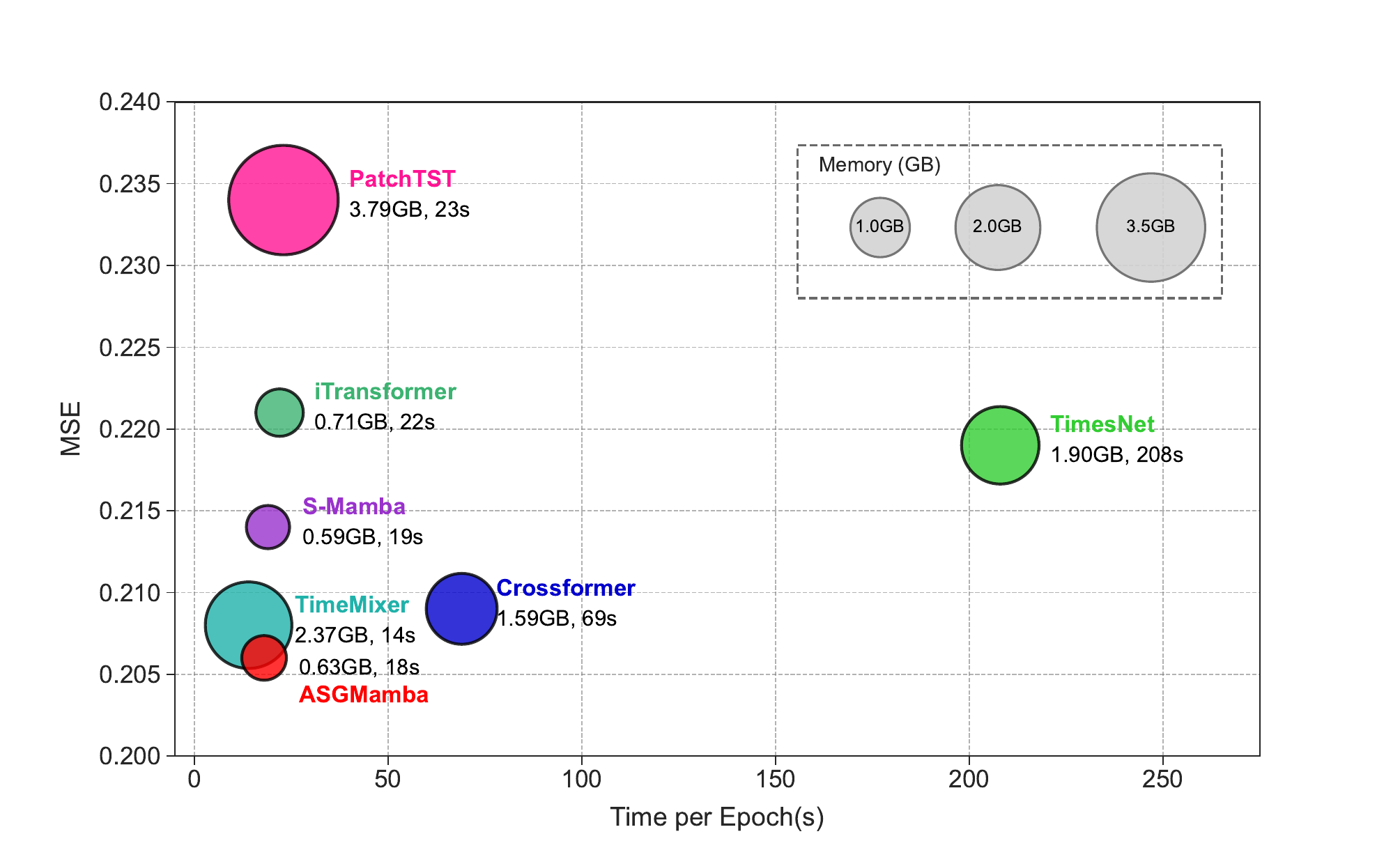} 
    \caption{Efficiency comparison on the weather dataset ($L=96, T=192$). The x-axis denotes training speed (s/epoch), the y-axis denotes MSE, and the bubble area represents GPU memory usage. ASGMamba (Red) achieves the optimal trade-off.}
    \label{fig:efficiency_comparison1}
\end{figure}

\begin{figure}[htbp]
    \centering
    \includegraphics[width=0.95\columnwidth]{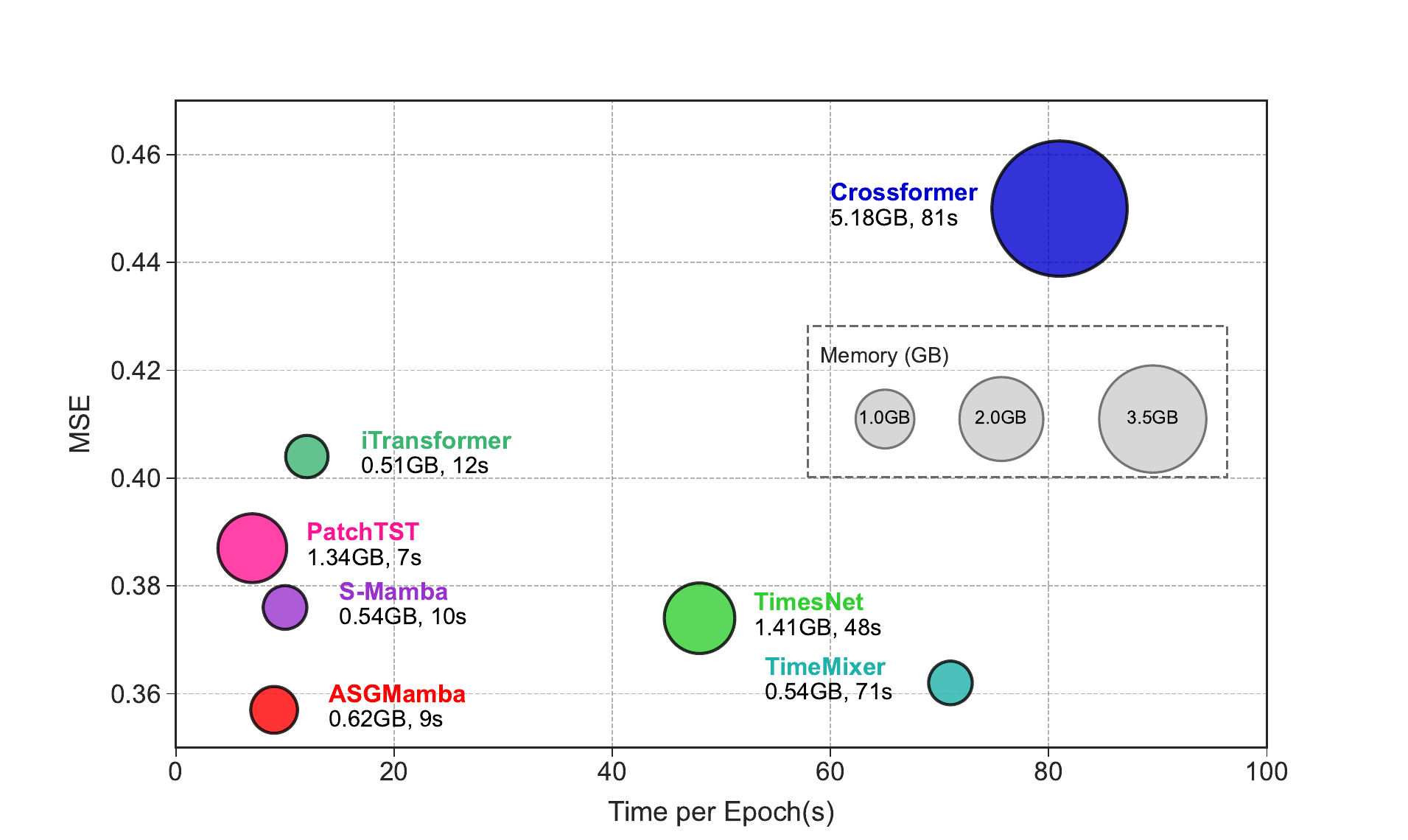} 
    \caption{Efficiency comparison on the ETTm1 dataset ($L=96, T=192$). The x-axis denotes training speed (s/epoch), the y-axis denotes MSE, and the bubble area represents GPU memory usage. ASGMamba (Red) achieves the optimal trade-off.}
    \label{fig:efficiency_comparison2}
\end{figure}

\subsection{Efficiency Analysis}

We evaluate the practical utility of forecasting models by analyzing the trade-off between predictive accuracy and computational cost. In Fig. \ref{fig:efficiency_comparison1} and Fig. \ref{fig:efficiency_comparison2}, we plot a high-dimensional performance space where x-axis denotes training speed, y-axis denotes forecasting error (MSE), and the bubble size denotes GPU memory footprint. Ideally, a model should be located in the bottom-left corner (Pareto frontier). Since we are interested in lightweight linear models are ruled out, high-capacity models are plotted. They are capable of modeling non-linear dynamics. ASGMamba (red bubble) is always able to achieve the best performance with least memory footprint.
Unlike Transformer-based architectures, ASGMamba shows a clear advantage due to the high memory cost of quadratic attention complexity $\mathcal{O}(L^2)$. For example, on Weather dataset, Crossformer needs 5.18GB memory and 81s/epoch due to its two-stage attention mechanism. While ASGMamba only needs 0.62GB memory (reduced by $8\times$) and 9s/epoch (9$\times$ speedup) while achieving lower MSE. The linear-complexity spectral gating indeed removes the redundancy existing in deep attention networks.
In addition, ASGMamba is significantly faster/finer than efficient CNN/MLP-based baselines. ASGMamba is approximately $4\times$ faster than TimesNet due to the heavy 2D convolutions used in TimesNet for temporal modeling. Unlike TimeMixer, average pooling will drop the high-frequency details of the signal. While our orthogonal wavelet decomposition model preserves the details of the signal. Therefore, ASGMamba achieves higher accuracy without latency overhead, making it more applicable to real-world scenarios with limited resources.

\section{Conclusion}
\label{sec:con}
In this work, we introduce ASGMamba, a forecasting framework that addresses the short-state-capacity issue of linear SSMs and the challenging statistical properties of real-world time series spectral characteristics.
We lift this limitation by designing an Adaptive Spectral Gating mechanism that lets the state evolve in a spectral-conditioned manner, enabling the filtering of broadband noise injected at the input stage, while avoiding the associated cost of performing global frequency transformations.
The proposed architecture maintains the conservative $\mathcal{O}(L)$ scaling of its Mamba backbone while keeping enhancing a great robustness to volatile dynamics.

Empirical evaluations across nine diverse benchmarks demonstrate that ASGMamba achieves a Pareto-optimal trade-off: it matches or exceeds the predictive accuracy of computationally intensive Transformer architectures while delivering high inference throughput and a minimal memory footprint. 
These attributes establish ASGMamba as a scalable solution particularly well-suited for high-frequency forecasting tasks in resource-constrained or latency-sensitive computational environments.

Future investigations will focus on extending this spectral-aware state space formulation to natively accommodate irregularly sampled or asynchronous time series. Furthermore, given its linear efficiency, we intend to explore ASGMamba as a token-efficient backbone for large-scale pre-trained time series foundation models.

\backmatter

\bmhead*{Acknowledgments}
This work is supported by the National Natural Science Foundation of China (No. 62372366).

\bmhead*{Data availability}
 Data is available on request.
 
\section*{Declarations}
\textbf{Conflict of interest} The authors declare that they have no known competing financial interests or personal relationships that could have appeared to influence the work reported in this paper.


\bibliography{ASGMamba}

\end{document}